\PassOptionsToPackage{table}{xcolor}

\documentclass[letterpaper]{article}
\usepackage[preprint]{aaai2027}
\usepackage[hyphens]{url}
\usepackage{graphicx}
\usepackage{natbib}
\usepackage{caption}
\usepackage{amsmath}
\usepackage{amssymb}
\usepackage{booktabs}
\usepackage{multirow}
\definecolor{mygray}{gray}{.94}

\title{LogiShot: Logically Coherent Cross-Shot Video Generation}

\author{
    Shuai Guo\textsuperscript{\rm 1},
    Yuhang Yang\textsuperscript{\rm 1},
    Zeyu Zhang\textsuperscript{\rm 1},
    Pengfei Yu\textsuperscript{\rm 2},\\
    Wei Zhai\textsuperscript{\rm 1},
    Yang Cao\textsuperscript{\rm 1},
    Zheng-Jun Zha\textsuperscript{\rm 1}
}
\affiliations{
    \textsuperscript{\rm 1}University of Science and Technology of China\\
    \textsuperscript{\rm 2}Li Auto Inc.
}

\begin{document}
\maketitle

\begin{abstract}
Generating cross-shot videos that are logically connected is essential for content creation. Currently, most cross-shot video-generation workflows, such as short-drama production, still rely on isolated textual scripts or explicit reference images to specify the generated content. Consequently, when user instructions are underspecified or ambiguous, a generated clip may appear visually plausible on its own but fail to align with the overall narrative, leading to disjointed content. We argue that achieving cross-shot logical coherence in video generation requires establishing logical connections across shots and maintaining visual consistency. To this end, we propose LogiShot, which incorporates information through two complementary paths: 1) LogiShot jointly encodes the context video and other conditioning signals, yielding dense multimodal cues that provide visual-semantic evidence for cross-shot generation; 2) the model maintains a visual memory of the context video throughout generation to preserve visual consistency across shots. Additionally, we construct a dataset with $110$K samples and a dedicated benchmark for evaluating cross-shot logical coherence. Experiments demonstrate that LogiShot consistently outperforms existing baselines in terms of logical coherence across multiple shots. Model and data will be made publicly available.
\end{abstract}

\begin{figure*}[t]
  \centering
  \includegraphics[width=\textwidth]{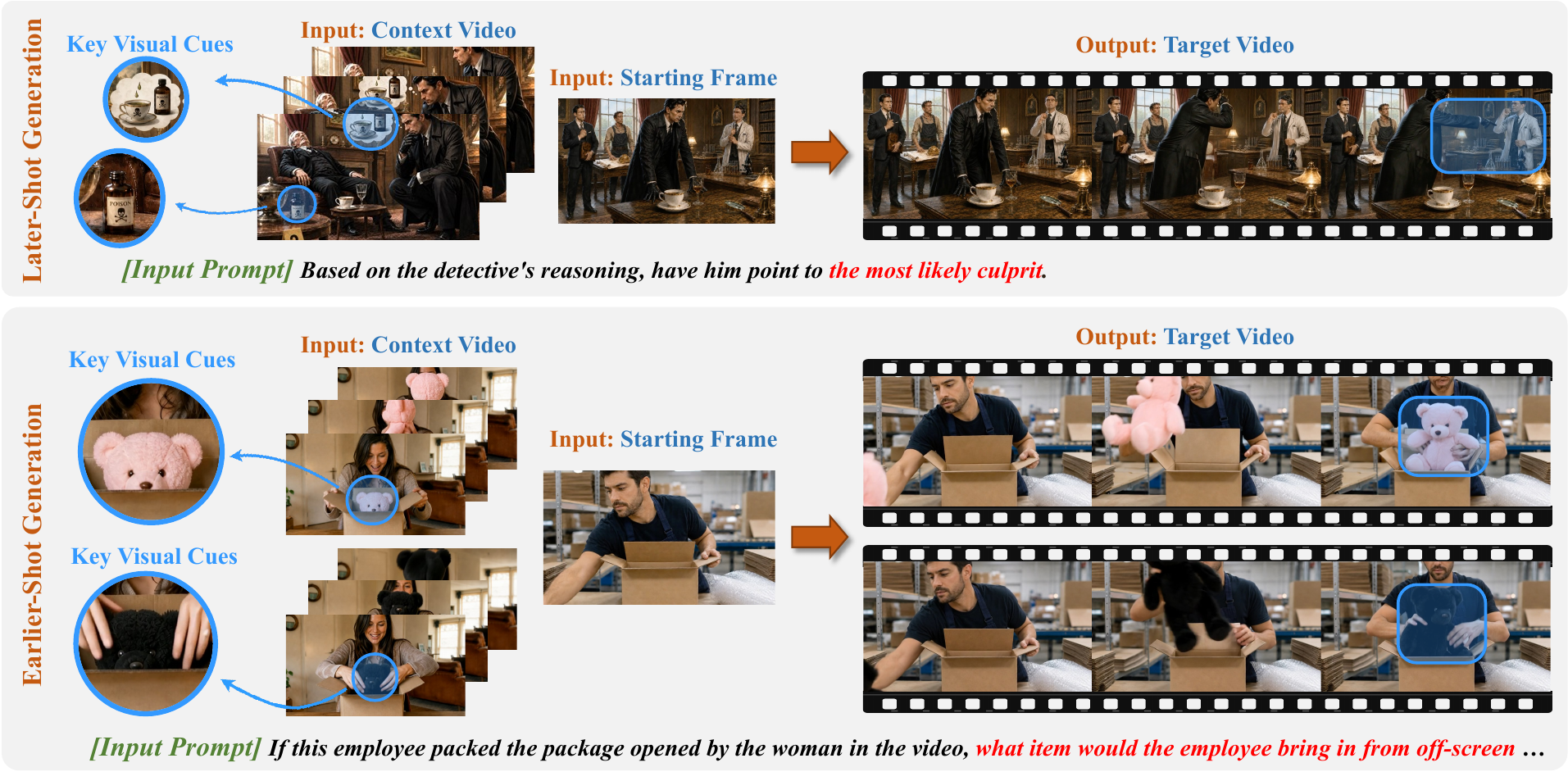}
  \caption{Logically coherent cross-shot generation.
  Given the context video, the input prompt, and a starting frame, our
  model generates a logically coherent video. In the
  top example, it generates a video in which the detective points to
  the likely culprit, following the causal relation implied by the
  context video. In the bottom example, it generates a video in which
  an employee brings in the same toy shown in the context video,
  preserving visual consistency across shots.}
  \label{fig:teaser}
\end{figure*}

\section{Introduction}
\label{sec:intro}
Video content creation often involves merging multiple segments to
form a coherent narrative. The events depicted in these segments are
not independent but are often linked by logical relations (e.g.,
causal or parallel). Maintaining logical coherence across segments
during the creative process is therefore essential for such downstream applications.

Recent video generation models can produce high-fidelity individual
clips~\cite{wan2025wan,kong2024hunyuanvideo,yang2025cogvideox,
polyak2024movie,teng2025magi,zeng2024dawn,yang2025sigman}. This independent clip-generation
paradigm requires users to specify how the content of each clip should
relate to other segments or shots. Thus, existing multi-shot and
story-generation methods produce multiple shots from detailed
user-provided descriptions, including per-shot captions or
scripts~\cite{an2026onestory,zhang2025storymem}. Some methods also use
explicit reference images to maintain logical coherence and visual
consistency across shots~\cite{yuan2025identity,liu2025phantom,
jiang2024videobooth,ma2024magic,yang2026gloria,zeng2026lpm,
chen2025dancetogether,li2024dispose}. However, these methods still depend
on isolated descriptions provided in advance. When user
instructions are underspecified or ambiguous, it becomes difficult to
determine how multiple video clips should be logically connected and
which details should remain consistent across them. Consequently, the
generated clips may appear visually plausible but fail to align with the overall narrative.

In this paper, we argue that achieving cross-shot logical coherence in
video generation requires establishing logical connections across shots
and maintaining visual consistency. To this end, we propose
\emph{LogiShot}, which takes the context video, the prompt instruction,
and a starting frame as inputs to generate logically related video
clips (Fig.~\ref{fig:teaser}). The key insight behind LogiShot is that
cross-shot video generation should account for the intended
logical relation between the target and context videos rather than
proceed in isolation. Accordingly, the context video provides visual
and semantic evidence for generating content that follows the intended
logical relation. The prompt instruction specifies the intended relation
between the context video and the target video. The starting frame
establishes the initial visual state of the generated video without
specifying how the event unfolds.

Realizing this insight hinges on a core challenge: how to effectively
fuse the three inputs so that cross-shot generation explicitly accounts
for the intended logical relation while preserving context-relevant
visual details. To address this challenge, we introduce two complementary
mechanisms: 1) Multimodal Cue Guidance (MCG) uses a frozen VLM to jointly
process the context video, the prompt instruction, and the starting
frame. MCG then augments the decoded target-event description with dense
multimodal cues that retain visual-semantic information not fully
captured by text alone. 2) Visual Memory (VM) retains context-video
latents accessible to the DiT throughout generation, allowing
target-video tokens to retrieve relevant visual details at every layer.
Moreover, we construct a dataset with 110K samples
and a dedicated benchmark for evaluating cross-shot logical coherence.
Our main contributions are summarized as follows:
\begin{itemize}
  \item We emphasize that establishing logical connections across shots and
  maintaining visual consistency are key to achieving cross-shot logical
  coherence, and propose \emph{LogiShot}, a video generation method that
  integrates the context video and other conditioning signals throughout
  generation to meet both requirements.
  \item We introduce two complementary mechanisms, Multimodal Cue Guidance (MCG) and Visual
  Memory (VM), which respectively use dense multimodal cues to guide
  the generation of logically related content and context-video
  latents to preserve visual consistency across shots.
  \item We construct a dataset with 110K cross-shot logically coherent samples
  and a dedicated benchmark for evaluating cross-shot logical
  coherence. Experiments show that LogiShot outperforms existing
  baselines across all metrics.
\end{itemize}

\noindent\textbf{Project page:}
\url{https://guoshuaigo.github.io/LogiShot/}.

\begin{figure*}[tp]
  \centering
  \includegraphics[width=0.95\linewidth]{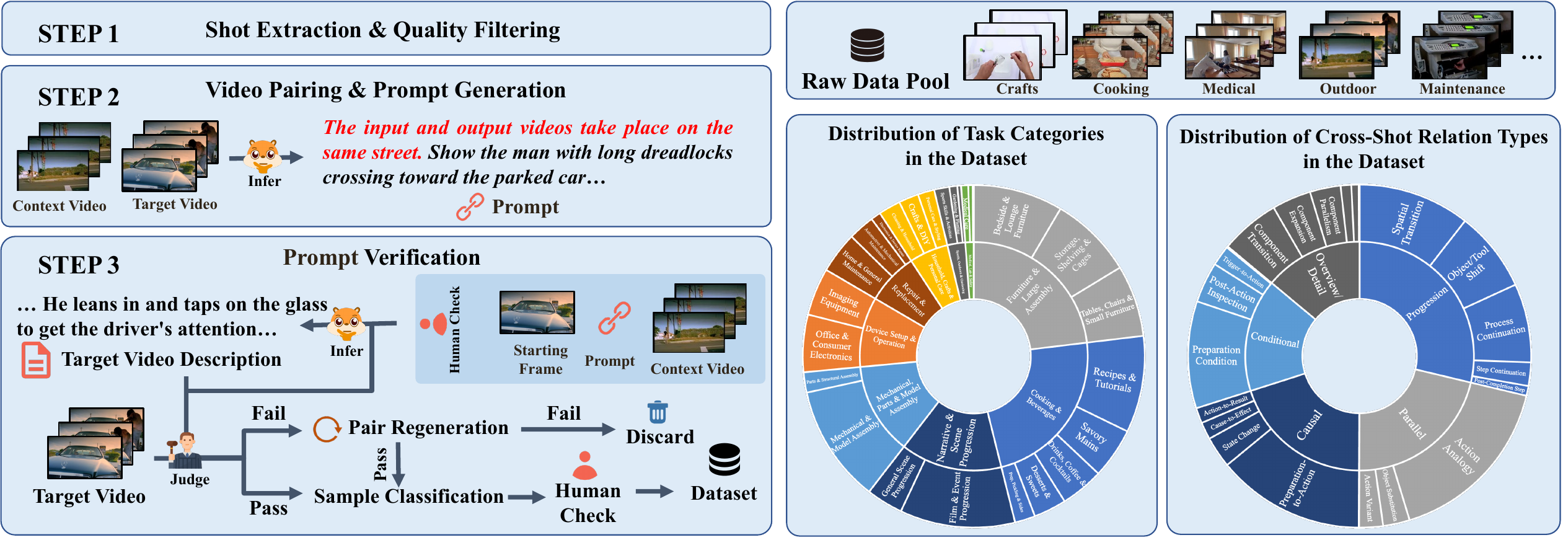}
  \caption{Data construction pipeline.
  Left: The three-stage pipeline that constructs the dataset
  from raw videos: Shot Extraction and Quality Filtering, Video
  Pairing and Prompt Generation, and Prompt Verification. Human audits
  are conducted after pair generation and prompt verification.
  Right:
  Distribution of the resulting dataset with 110K
  samples across task categories and cross-shot
  relation types.}
  \label{fig:data_pipeline}
\end{figure*}

\section{Related Work}
\label{sec:related}
\paragraph{Cross-Shot Video Generation.}
Cross-shot video generation includes multi-shot systems, which
organize long-form content into a sequence of planned shots, and
story-generation systems, which expand a story idea or script into
detailed shot descriptions and use storyboards or reference assets to
guide the generation of individual shots~\cite{long2024videostudio,lin2023videodirectorgpt,
zhuang2024vlogger,wu2025automated,zheng2024videogen,
xiao2025videoauteur}. In addition, to maintain visual consistency across shots,
they employ joint generation or cross-shot conditioning~\cite{meng2026holocine,guo2025long,
he2025cut2next,an2026onestory,zhang2026stage,luo2026shotstream}.
However, their generation process relies primarily on predefined
descriptions or references, making it difficult to determine how a
generated shot should be logically connected to the context video when
user instructions are underspecified or ambiguous. In contrast,
LogiShot uses the context video to guide the generation of content that
follows the intended logical relation while maintaining visual
consistency across shots.

\paragraph{Reasoning in Video Generation.}
Recent video reasoning models have shown strong capabilities in
analyzing causal and temporal relations among
events~\cite{feng2026video,fei2024video,min2024morevqa,
yang2023grounding,yang2024lemon,yang2024egochoir,shao2025great}.
Transferring these reasoning capabilities to video generation remains challenging
because video reasoning models typically express their predictions as discrete labels,
structured event representations, or textual
descriptions~\cite{yi2019clevrer,xiao2021next,wu2024star,
li2023intentqa,lei2020more,wang2026fostering,chen2024mecd,
ayyubi2025enter}. Recent approaches bridge this gap
through next-event prediction and reasoning-guided
generation~\cite{li2026happens,huang2025vchain,
shen2025counterfactual,spyrou2025causally,yu2025hero,han2026touch}.
For instance, VANS~\cite{cheng2026video} conditions its generator on a
predicted event caption and sampled VAE tokens. However, in such
frameworks, the VLM conveys its event-level interpretation to the
generator primarily through the decoded caption; contextual details
not verbalized in the caption are absent from the generator's semantic
condition.
Although unified video models connect MLLMs to diffusion generators
through continuous visual representations
~\cite{tan2025omni,wei2026univideo,luo2025univid},
they focus on general-purpose video understanding and generation
rather than explicitly using these representations to generate content
that is logically connected to the context video. In contrast,
LogiShot conditions the DiT on
both the decoded target-event description and dense multimodal cues
from the VLM, allowing context-specific visual-semantic
information beyond the decoded description to guide the generation of
logically related content.

\section{Data Construction}
\label{sec:data_construction}
Training LogiShot requires samples in which the target event can be
inferred only from the context video, the prompt instruction,
and the starting frame jointly---a format unavailable in existing public corpora.
Each sample includes these three inputs together with a
target video as supervision.
We construct this dataset through
a three-stage pipeline (Fig.~\ref{fig:data_pipeline}).
This pipeline extracts clips from public sources, generates candidate pairs,
and filters them to retain only those meeting this criterion.
Two human audits serve as independent quality gates at different stages
of the pipeline: one after pair generation and another after prompt
verification.

\paragraph{Shot Extraction, Video Pairing, and Prompt Generation.}
Raw videos come from public event-centric datasets~\cite{liang2024guide,chen2024mecd,liang2025videvent}.
Each source video is segmented into event-level clips via shot-boundary detection,
and a rule-based visual filter removes low-quality clips.
Each retained clip is then processed by a VLM to extract a structured description
of the actor, salient objects, action, and resulting state.
Two clips from the same source video are paired:
one serves as the context video and the other as the target video,
with the target's first frame designated as the starting frame.
Finally, each pair is annotated with a prompt instruction
that captures the logical relation connecting them without explicitly
detailing the target event.

\paragraph{Prompt Verification and Human Audits.}
A candidate pair is retained only if its target event is deducible
from the three inputs taken together, rather than from any subset of
them. To verify this, a VLM
is given the three inputs and predicts a target-event description. A judge then compares this
prediction with the ground-truth target video using a five-point rubric.
The rubric evaluates whether the prediction and target video depict the
same event and whether inferring that event requires the full context.
Pairs scoring below the acceptable threshold undergo pair regeneration
to resolve ambiguities or remove overly explicit cues and are then
re-evaluated. Those still below the threshold are discarded. Accepted
pairs are assigned a quality score by the judge.
To prevent data leakage across splits, we apply near-duplicate filtering and hold out
an evaluation subset at the raw-source-video level. The resulting dataset comprises
110K samples, with relation and task-category
distributions detailed in Fig.~\ref{fig:data_pipeline}.
A human audit of 1,000 pairs from the final dataset yields a
composite pass rate of 89.3\%. On a separate 500-pair validation set,
the prompt-verification judge agrees with the three-annotator consensus
in 91.6\% of cases.

\section{Method}
\label{sec:method}
\begin{figure*}[t!]
  \centering
  \includegraphics[width=0.95\linewidth]{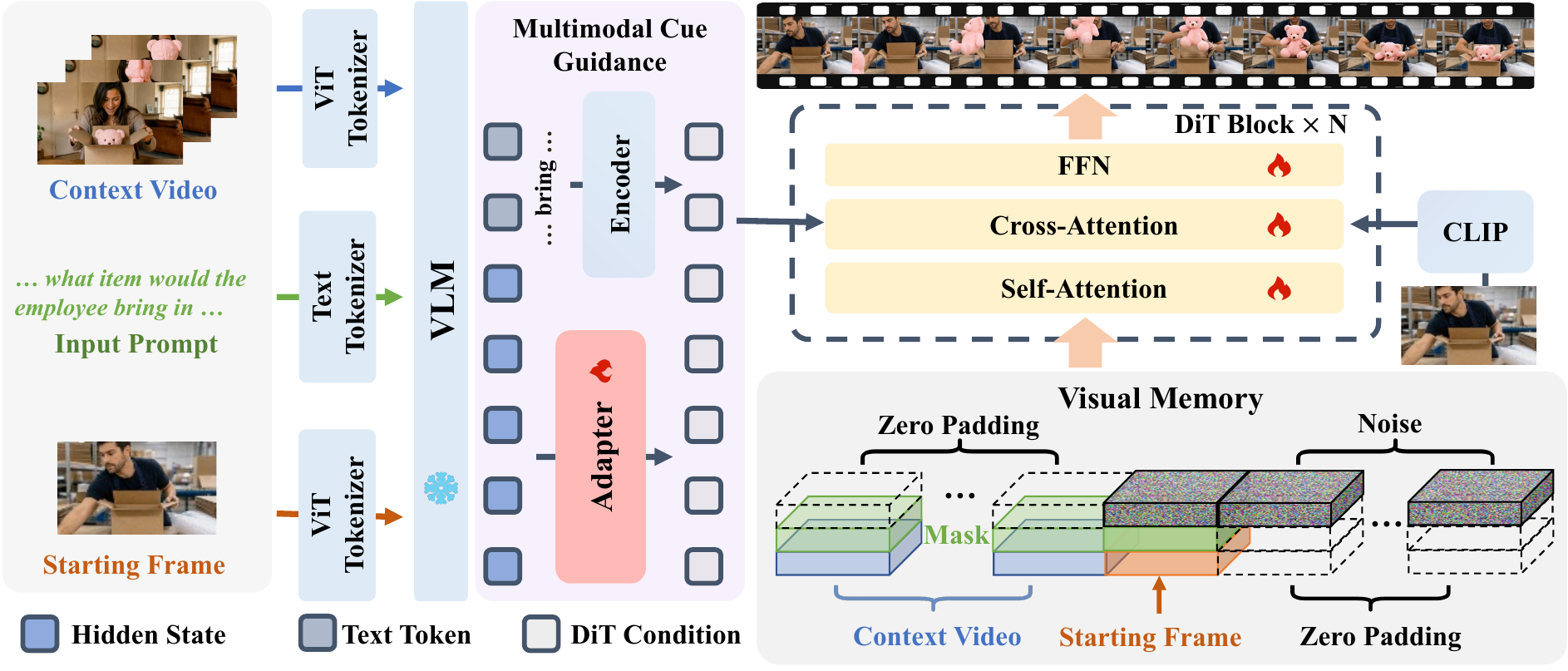}
  \caption{LogiShot architecture.
  Multimodal Cue Guidance (MCG; Sec.~\ref{sec:event_inference})
  augments a target-event description with dense multimodal cues,
  while Visual Memory
  (VM; Sec.~\ref{sec:entity_cond}) allows target-video tokens to
  retrieve context-video information through self-attention. The
  starting frame provides the initial visual state for the generated
  video.}
  \label{fig:method}
\end{figure*}
Given the context video $V\in\mathbb{R}^{T_c\times 3\times H\times W}$,
the prompt instruction $p$, and a starting frame
$I_0\in\mathbb{R}^{3\times H\times W}$, LogiShot generates a target video
$\hat V\in\mathbb{R}^{T_t\times 3\times H\times W}$, where $T_c$ and
$T_t$ denote the context and target frame counts, and $H\times W$ is the
spatial resolution. The generated video
begins with $I_0$, is logically related to $V$ according to $p$, and
maintains visual consistency with $V$.

Achieving this hinges on a core challenge: how to effectively fuse
$V$, $p$, and $I_0$ so that the generated video both realizes the
intended relation and preserves context-relevant visual details. To
address this challenge, we introduce two complementary mechanisms that
fuse the inputs along different paths into a video DiT
backbone~\cite{peebles2023scalable} (Fig.~\ref{fig:method}). Our first mechanism, Multimodal Cue
Guidance (MCG), uses a frozen VLM to jointly process $V$, $p$, and $I_0$
and augments its decoded target-event description with dense
multimodal cues that retain visual-semantic information beyond text
alone. Our second mechanism, Visual Memory (VM), retains context-video
latents accessible to the DiT throughout generation, allowing target-video
tokens to retrieve relevant visual information from $V$ at every layer.
The starting frame $I_0$ serves as the fixed initial visual state of
the generated video. The rest of this section
describes MCG (Sec.~\ref{sec:event_inference}), VM and starting-frame
conditioning (Sec.~\ref{sec:entity_cond}), and the training and
inference procedures (Sec.~\ref{sec:training_inference}).

\subsection{Multimodal Cue Guidance}
\label{sec:event_inference}
$p$ indicates the intended logical relation but may leave the
corresponding target event underspecified. A frozen VLM $\Psi$ jointly
processes $V$, $p$, and $I_0$ and autoregressively decodes an explicit
target-event description $e$. This description provides explicit
semantic guidance, but may not fully capture the visual-semantic
information retained in $\Psi$'s internal computation. MCG therefore
augments the decoded description with dense multimodal cues extracted
from the same VLM forward pass.

We tokenize $e$ and encode it with the DiT's text encoder $\tau$ to obtain
text-conditioning tokens
$\tau(e)\in\mathbb{R}^{L_e\times d_t}$, where $L_e$ is the number of text tokens and $d_t{=}4096$.

From the same VLM forward pass over $(V,p,I_0)$, we extract $\Psi$'s
final-layer hidden states,
$\mathbf{H}\in\mathbb{R}^{L_H\times d_v}$, where $L_H$ is the sequence length and $d_v{=}3584$, and map
them to the DiT's text dimension using a learned projector
$P:\mathbb{R}^{d_v}\!\to\!\mathbb{R}^{d_t}$, implemented as a
two-layer MLP with GELU. We refer to the projected hidden states
$P(\mathbf{H})$ as dense multimodal cues and concatenate them with the
text-conditioning tokens:
\begin{equation}
  \mathbf{c}
  = \big[\,P(\mathbf{H});\;\tau(e)\,\big]
  \in \mathbb{R}^{(L_H+L_e)\times d_t}.
\end{equation}
The resulting $\mathbf{c}$ is the complete conditioning sequence for
the DiT. Each cross-attention block attends to $\mathbf{c}$, allowing the
decoded description and dense multimodal cues to jointly
guide target-video generation.

\begin{figure*}[t]
  \centering
  \includegraphics[width=0.98\linewidth]{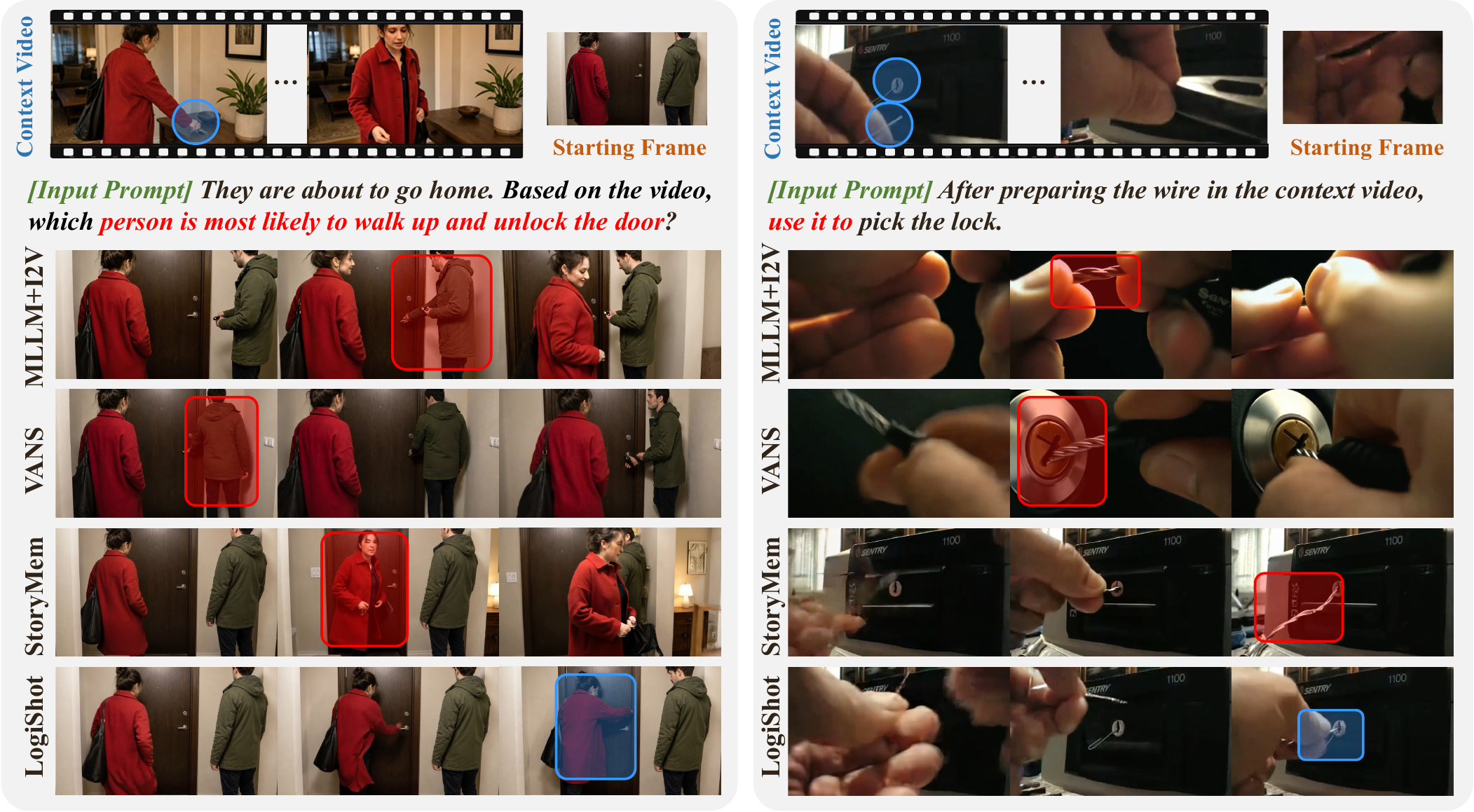}
  \caption{Qualitative comparison against baselines.
  Red boxes mark baseline failures. On the left, the baselines send
  the wrong actor to unlock the door or misread the spatial relation
  between the context video and the target video; on the right, they
  replace or distort the wire or the lock from the context video.
  Blue boxes mark LogiShot's correct behavior: the key holder
  unlocks the door (left), and the same wire is used to pick the same lock
  (right).}
  \label{fig:qualitative}
\end{figure*}

\subsection{Visual Memory}
\label{sec:entity_cond}
To maintain visual consistency, VM retains a prefix of context-video
latent slots that remains accessible to the DiT throughout generation.
Through self-attention in every transformer block, target-video tokens
can retrieve relevant visual information from this prefix. The starting
frame is incorporated separately through latent and CLIP conditioning
paths to establish the initial visual state.

\paragraph{Latent-Slot Construction.}
We cast frames sampled from the context video and the starting
frame as temporal conditioning slots in the same latent space as the
target video. Each slot is formed by concatenating a presence mask with
a VAE latent along the channel dimension:
\begin{equation}
  S_i = \big[\,M_i;\;\mathcal{E}(x_i)\,\big]
  \in \mathbb{R}^{(d_M+d_Z)\times h\times w},
\end{equation}
where $\mathcal{E}$ denotes the video VAE and $M_i$ is a presence
mask that is nonzero only for the provided conditioning slots. We
uniformly sample $K$ frames from $V$ to form a context-video prefix.
The latent of $I_0$ is placed at the first target-video slot, while the
remaining target-video slots are zero-filled in both their masks and
latents:
\begin{equation}
  S=\big[\,S^{v}_1,\,\ldots,\,S^{v}_K,\;S^{0},\;\mathbf{0},\,\ldots,\,\mathbf{0}\,\big].
\end{equation}
The target-video noise sequence is left-padded with $K$ empty slots to
align it with $S$. The aligned noise and conditioning sequences are
then concatenated along the channel dimension before patch embedding.

\paragraph{Attention over Visual Memory.}
The starting frame is additionally encoded by CLIP, whose features
$\Phi(I_0)$ are injected through a parallel cross-attention path
alongside $\mathbf{c}$. Within the DiT, the context-video prefix, the
starting-frame slot, and the noisy target-video slots interact through
unified bidirectional self-attention. We apply 3D rotary positional
encoding over the $(\text{frame},\text{height},\text{width})$ axes to
preserve their spatiotemporal positions within the joint sequence. Through self-attention,
target-video tokens can retrieve relevant visual information from the
context-video prefix in every transformer block. The prefix therefore
serves as a visual memory rather than a set of frames to be copied
directly. Meanwhile, the starting-frame latent and CLIP features provide
the initial visual state for the generated video without determining
how the subsequent event unfolds.

\begin{table}[t]
  \centering
  \setlength{\tabcolsep}{1mm}
  \renewcommand{\arraystretch}{1.05}
  {\footnotesize
  \begin{tabular}{l|ccc|ccc}
    \toprule
    & \multicolumn{3}{c|}{Logical Correctness\,$\uparrow$}
    & \multicolumn{3}{c}{Visual Consistency\,$\uparrow$}\\
    \cmidrule(lr){2-4}\cmidrule(lr){5-7}
    Method & EvMt & RelMt & Comp
           & PartMt & RoleMt & StMt\\
    \midrule
    MLLM+I2V
      & 0.798 & 0.840 & 0.791
      & 0.853 & 0.850 & 0.835 \\
    VANS
      & 0.825 & 0.858 & 0.819
      & 0.876 & 0.870 & 0.867 \\
    StoryMem
      & 0.790 & 0.790 & 0.789
      & 0.893 & 0.892 & 0.891 \\
    \rowcolor{mygray}
    \textbf{LogiShot}
      & \textbf{0.901} & \textbf{0.939} & \textbf{0.887}
      & \textbf{0.938} & \textbf{0.931} & \textbf{0.931}
      \\
    \bottomrule
  \end{tabular}}
  \caption{Main results on the $800$-sample held-out evaluation set.
  Logical Correctness (LC) and Visual Consistency (VC) are reported
  separately through the three sub-metrics defined in
  Sec.~\ref{sec:benchmark}; higher is better.}
  \label{tab:main}
\end{table}

\subsection{Training Objective and Inference}
\label{sec:training_inference}
We train the denoiser using a rectified-flow objective. Given the
target-video latents $z_0=\mathcal{E}(V^{\star})$ of the ground-truth
target video $V^{\star}$ and a noise level $\sigma\in[0,1]$ drawn from a
shifted flow-matching schedule, we formulate the noisy latents as
$z_\sigma=(1-\sigma)z_0+\sigma\epsilon$, where
$\epsilon\sim\mathcal N(\mathbf 0,\mathbf I)$, and regress the corresponding
velocity field at that noise level:
\begin{equation}
  \mathcal{L}
  = \mathbb{E}_{\sigma,\,z_0,\,\epsilon}
  \big\|\,D_\theta\!\big(z_\sigma, \sigma \mid \mathbf{c}, \Phi(I_0), S\big)
  - u_\sigma\,\big\|_2^2 ,
\end{equation}
where the target velocity is $u_\sigma=\epsilon-z_0$. Since the
context-video prefix and the starting-frame slot serve as conditioning
inputs rather than reconstruction targets, the loss is evaluated only
on the target-video slots.

During inference, we first compute the conditioning representations
$\mathbf{c}$, $\Phi(I_0)$, and $S$ from the inputs $(V, p, I_0)$. The
denoiser then iteratively denoises the target-video
slots under classifier-free guidance to generate the final video $\hat V$.

\begin{figure*}[!t]
  \centering
  \includegraphics[width=0.98\linewidth]{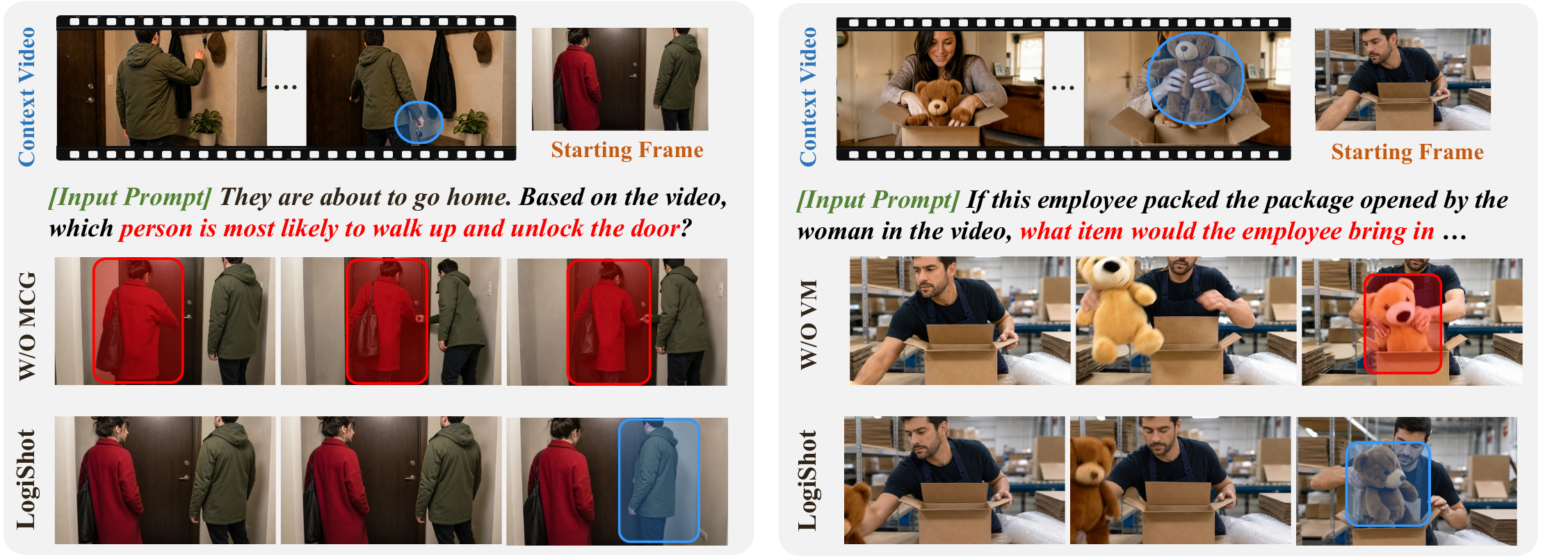}
\caption{Qualitative ablation of MCG and VM.
  The \emph{W/O MCG} variant assigns the unlock action to the wrong
  actor (left), while the \emph{W/O VM} variant fails to preserve the
  teddy bear's appearance (right).}
  \label{fig:ablation_qual}
\end{figure*}

\section{Experiments}
\label{sec:experiments}
\paragraph{Implementation.}
We initialize our generator with a $14$B-parameter image-to-video DiT
and update only its denoiser and conditioning projections during training.
All videos are generated as $49$-frame clips at $16$~fps with a resolution
of $480{\times}832$. Additional implementation details, including optimizer
settings, frozen encoders, and data preprocessing steps, are provided in
Appendix~\ref{sec:supp_impl}.

\subsection{Benchmark and Evaluation Protocol}
\label{sec:benchmark}
\paragraph{Evaluation Protocol.}
We evaluate all methods on the $800$-sample held-out evaluation set,
generating one video per sample. Anonymized outputs are scored by a
VLM judge along two primary dimensions, each computed as the average of
three sub-metrics. \emph{Logical Correctness} (LC) evaluates whether the
generated video depicts the event implied by the inputs, follows the
intended logical relation to the context video, and completes the event.
It comprises Event Match (EvMt), Relation Match (RelMt), and Completion
(Comp). \emph{Visual Consistency} (VC) evaluates whether context-relevant
participants, their roles, and their states remain consistent between the
context and generated videos. It comprises Participant Match (PartMt), Role
Match (RoleMt), and State Match (StMt). The evaluation judge is validated
against human ratings on a $150$-case validation subset, showing strong
agreement, with a
quadratic-weighted Cohen's $\kappa$ of $0.84$ for LC. Across the three
baseline comparisons, all $18$ sub-metric differences are statistically
significant under paired randomization tests after Holm correction
($p<0.01$ in all cases).
\begin{table}[tp]
  \centering
  \setlength{\tabcolsep}{1mm}
  \renewcommand{\arraystretch}{1.05}
  {\footnotesize
  \begin{tabular}{cc|ccc|ccc}
    \toprule
    \multicolumn{2}{c|}{Mechanism}
    & \multicolumn{3}{c|}{Logical Correctness\,$\uparrow$}
    & \multicolumn{3}{c}{Visual Consistency\,$\uparrow$}\\
    \cmidrule(lr){1-2}\cmidrule(lr){3-5}\cmidrule(lr){6-8}
    MCG & VM
       & EvMt & RelMt & Comp
       & PartMt & RoleMt & StMt\\
    \midrule
               &            & 0.790 & 0.804 & 0.791
                            & 0.850 & 0.845 & 0.830 \\
    \checkmark &            & 0.833 & 0.849 & 0.828
                            & 0.882 & 0.879 & 0.875 \\
               & \checkmark & 0.792 & 0.810 & 0.786
                            & 0.900 & 0.900 & 0.894 \\
    \rowcolor{mygray}
    \checkmark & \checkmark
      & \textbf{0.901} & \textbf{0.939} & \textbf{0.887}
      & \textbf{0.938} & \textbf{0.931} & \textbf{0.931} \\
    \bottomrule
  \end{tabular}}
  \caption{Ablation of MCG and VM.
  The decoded target-event description and starting frame are retained
  in all variants. The reported sub-metrics follow Tab.~\ref{tab:main};
  higher is better.}
  \label{tab:ablation}
\end{table}

\paragraph{Baselines.}
We compare LogiShot against three baselines, adapting each to use the same inputs: the context video, the prompt instruction, and a starting frame.
The baselines cover the main design axes for cross-shot generation:
no context conditioning, reasoning-generated text conditioning, and
frame-memory conditioning.
For every baseline, the VLM jointly processes all three inputs when
producing the target-event description.
\textbf{MLLM+I2V} feeds the resulting description and starting frame
to a standard image-to-video generator.
\textbf{VANS}~\cite{cheng2026video} conditions its generator on the
VLM-produced description and sampled context-video VAE tokens; we
adapt it to use the starting frame and match our parameter budget for
a fair comparison.
\textbf{StoryMem}~\cite{zhang2025storymem} stores selected context-video
frames in its memory bank and uses the VLM-produced target-event
description together with the starting frame for generation.

\paragraph{User Study.}
To complement the automated evaluation with human judgments, we conduct a
blind pairwise A/B study with $20$ annotators, with pairs drawn from the
same evaluation set. Each annotator evaluates $90$ anonymized video pairs
($30$ pairs comparing LogiShot against each of the three baselines). For
each pair, the annotator provides separate judgments for Logical Correctness,
Visual Consistency, and overall preference, with ties scored as $0.5$.

\subsection{Main Comparison}
\label{sec:main_comparison}
Tab.~\ref{tab:main} reports the quantitative results. LogiShot outperforms
all three baselines on every sub-metric, yielding gains of
$0.075$--$0.119$ in Logical Correctness and $0.041$--$0.087$ in Visual
Consistency. The blind pairwise user study further corroborates this
advantage: human raters prefer LogiShot over each baseline in
$73.0$--$76.3\%$ of pairwise comparisons for overall preference.
Detailed results are provided in Appendix~\ref{sec:supp_user_study}.

Together, these results demonstrate LogiShot's advantage in both
Logical Correctness and Visual Consistency. In particular,
along the text-conditioning axis, LogiShot improves Relation Match by $0.082$--$0.100$ over MLLM+I2V
and VANS, suggesting that conditioning on MCG's dense multimodal cues
provides the generator with relation-relevant information
that may not be captured in the decoded event description when the prompt
instruction is underspecified. Compared with StoryMem, LogiShot
improves Logical Correctness by $0.119$ while further increasing Visual
Consistency by $0.041$, highlighting the benefit of combining MCG with
VM. These quantitative trends are also reflected in
Fig.~\ref{fig:qualitative}. In these examples, the baselines assign
actions to the wrong actor or alter context-relevant objects, whereas
LogiShot generates the relation-consistent event while preserving the
relevant participants and object appearances.

\subsection{Ablations}
\label{sec:ablations}
Tab.~\ref{tab:ablation} evaluates the contributions of MCG and VM.
In this ablation, W/O MCG retains the decoded target-event description
but removes the dense multimodal cues $P(\mathbf{H})$. W/O VM removes
only the context-video latent prefix, while the starting-frame latent
and CLIP features remain available in all settings. The first row
disables both MCG and VM, serving as the base
variant conditioned only on the decoded description and the starting
frame. Enabling MCG alone improves Logical
Correctness by $+0.042$ (with the most pronounced gain in Relation Match,
$\Delta{=}0.045$), showing the benefit of the additional visual-semantic
information retained by the dense cues. Conversely, enabling
VM alone boosts Visual Consistency by $+0.056$, a gain consistent across
all three sub-metrics ($\Delta{\in}[0.050,\,0.064]$). Enabling both
mechanisms yields the highest overall performance ($+0.114$ LC and
$+0.092$ VC over the baseline), demonstrating that they are highly
complementary rather than redundant. Fig.~\ref{fig:ablation_qual}
visualizes the failure modes when either mechanism is disabled.

\subsection{Analysis}
\label{sec:analysis}
\paragraph{Performance Analysis.}
\begin{table}[tp]
  \centering
  \setlength{\tabcolsep}{1mm}
  \renewcommand{\arraystretch}{1.05}
  {\footnotesize
  \begin{tabular}{l|cccc}
    \toprule
    Method & Subj.\,$\uparrow$ & Mot.\,$\uparrow$
           & Tmp.\,$\uparrow$ & Avg.\,$\uparrow$ \\
    \midrule
    MLLM+I2V   & 0.892 & 0.977 & 0.963 & 0.944 \\
    VANS~\cite{cheng2026video}  & 0.889 & 0.976 & 0.961 & 0.942 \\
    StoryMem~\cite{zhang2025storymem}  & 0.886 & 0.978 & 0.963 & 0.942 \\
    \rowcolor{mygray}
    \textbf{LogiShot}
      & \textbf{0.926} & \textbf{0.985}
      & \textbf{0.976} & \textbf{0.962} \\
    \bottomrule
  \end{tabular}}
  \caption{Visual-quality diagnostics on the $800$-sample evaluation set.}
  \label{tab:vbench}
\end{table}
\begin{figure}[tp]
  \centering
  \includegraphics[width=\linewidth]{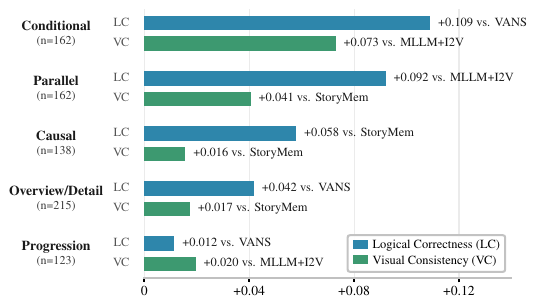}
  \caption{Per-relation-type gap. The gap between LogiShot and the
  strongest baseline for each relation type and evaluation dimension. Bars
  report Logical Correctness (LC) and Visual Consistency (VC) separately;
  the corresponding strongest baseline is annotated on each bar, and the
  sample count for each relation type is shown on the left.}
  \label{fig:relation_type}
\end{figure}
We conduct two diagnostic analyses to assess LogiShot's visual
quality and determine whether its gains extend beyond temporal
continuation.
First, we compute three VBench diagnostics~\cite{huang2024vbench,yang2025videogen} on
the $800$ samples: subject consistency (Subj.), motion smoothness (Mot.),
and temporal flickering (Tmp.); Avg.\ is their unweighted mean. LogiShot
achieves the highest score on every diagnostic, indicating that our
gains do not come at the expense of perceptual quality
(Tab.~\ref{tab:vbench}).
Second, the relation-wise breakdown (Fig.~\ref{fig:relation_type}) shows
that LogiShot's Logical Correctness advantage is largest on \emph{Conditional}
($+0.109$) and \emph{Parallel} ($+0.092$) relations.
These relations link context and target videos logically rather than
chronologically, requiring robust cross-shot reasoning.
\begin{figure}[tp]
  \centering
  \includegraphics[width=\linewidth]{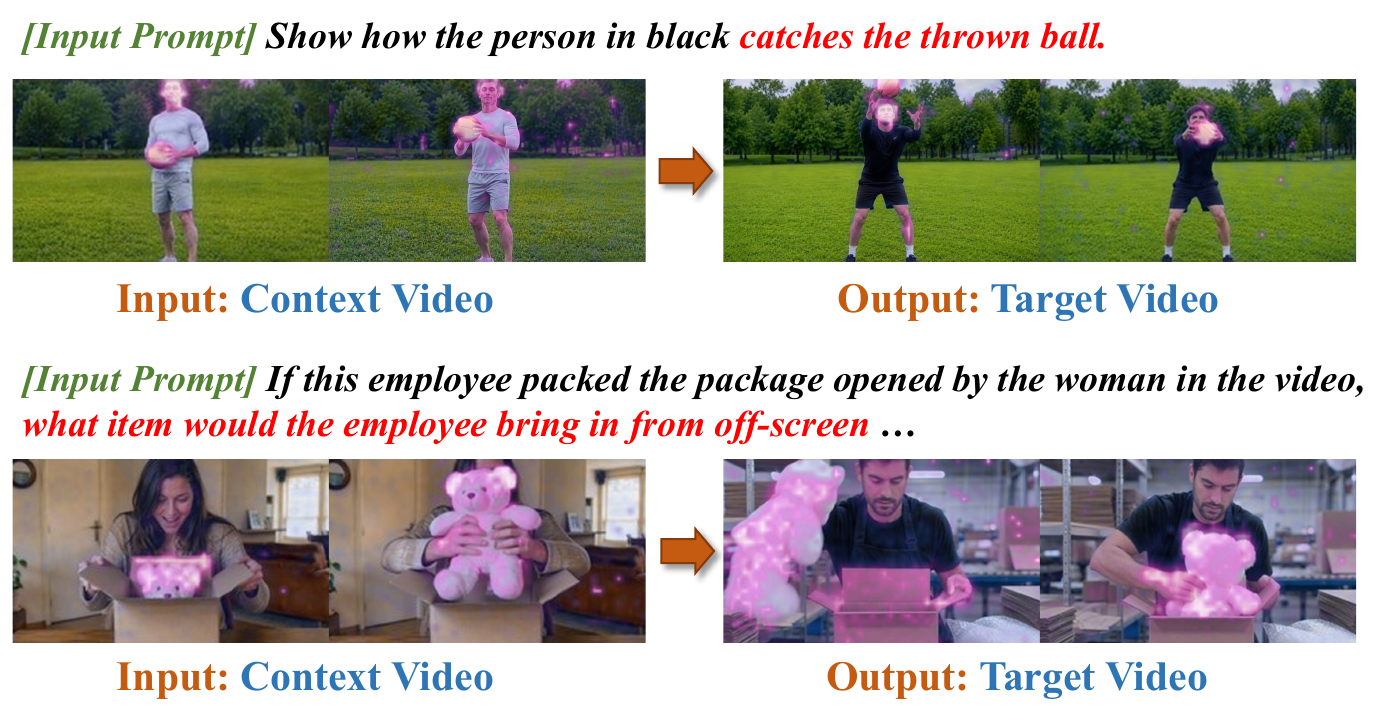}
  \caption{Attention over the context-video prefix.
  Pink overlays highlight the context-video regions receiving the highest
  attention weights from target-video tokens.}
  \label{fig:analysis_cases}
\end{figure}

\paragraph{Mechanism Analysis.}
To understand how LogiShot leverages the context video during generation, we visualize the
self-attention weights from target-video tokens to the context-video prefix
latents (Fig.~\ref{fig:analysis_cases}). In both cases, the peak
attention lands on the participant that carries the cross-shot relation:
the thrown ball in the top row, and the teddy bear that is transferred
into the target video in the bottom row. Quantitatively, during early
denoising, $56.9$--$68.8\%$ of the top-$1$ attention peaks from
target-video tokens fall within the top $5\%$ most-attended spatial regions
of the context-video across the two analyzed cases. This concentration
decreases during the final denoising steps. This attention pattern suggests
that the model first focuses on
relation-relevant context regions to establish coherence and then
distributes its attention more broadly during fine-grained generation. Together,
these cases provide qualitative evidence that LogiShot uses the context
video as a dense visual memory rather than relying solely on abstract text
summaries.

\section{Conclusion}
\label{sec:conclusion}
In this work, we argue that achieving cross-shot logical coherence in video
generation requires establishing logical connections across shots and maintaining
visual consistency. To achieve this, we propose \emph{LogiShot}, which introduces
two complementary mechanisms, Multimodal Cue Guidance (MCG) and Visual Memory (VM), into a video DiT.
MCG augments a decoded target-event
description with dense multimodal cues to guide the generation of
logically related content, while VM maintains context-video
latents as a visual memory throughout generation to preserve visual
consistency. We further construct a dataset with 110K
samples and a dedicated benchmark for evaluating cross-shot logical
coherence. Experiments demonstrate that LogiShot
outperforms existing baselines across all metrics. LogiShot represents a
step toward generating logically related video clips with less reliance
on exhaustive per-shot descriptions.

\bibliography{main}

\begin{thebibliography}{67}
\providecommand{\natexlab}[1]{#1}

\bibitem[{An et~al.(2026)An, Jia, Qiu, Zhou, Huang, Liu, Ren, Kahatapitiya,
  Liu, He et~al.}]{an2026onestory}
An, Z.; Jia, M.; Qiu, H.; Zhou, Z.; Huang, X.; Liu, Z.; Ren, W.; Kahatapitiya,
  K.; Liu, D.; He, S.; et~al. 2026.
\newblock Onestory: Coherent multi-shot video generation with adaptive memory.
\newblock In \emph{Proceedings of the IEEE/CVF conference on computer vision
  and pattern recognition}, 16173--16184.

\bibitem[{Ayyubi et~al.(2025)Ayyubi, Liu, Asgarov, Hakim, Sarker, Wang, Tang,
  Alomari, Atabuzzaman, Lin et~al.}]{ayyubi2025enter}
Ayyubi, H.; Liu, J.; Asgarov, A.; Hakim, Z. I.~A.; Sarker, N.~H.; Wang, Z.;
  Tang, C.-W.; Alomari, H.; Atabuzzaman, M.; Lin, X.; et~al. 2025.
\newblock Enter: Event based interpretable reasoning for videoqa.
\newblock \emph{arXiv preprint arXiv:2501.14194}.

\bibitem[{Bai et~al.(2025)Bai, Chen, Liu, Wang, Ge, Song, Dang, Wang, Wang,
  Tang et~al.}]{bai2025qwen25vltechnicalreport}
Bai, S.; Chen, K.; Liu, X.; Wang, J.; Ge, W.; Song, S.; Dang, K.; Wang, P.;
  Wang, S.; Tang, J.; et~al. 2025.
\newblock {Qwen2.5-VL} Technical Report.
\newblock \emph{arXiv preprint arXiv:2502.13923}.

\bibitem[{Caron et~al.(2021)Caron, Touvron, Misra, J{\'e}gou, Mairal,
  Bojanowski, and Joulin}]{caron2021emerging}
Caron, M.; Touvron, H.; Misra, I.; J{\'e}gou, H.; Mairal, J.; Bojanowski, P.;
  and Joulin, A. 2021.
\newblock Emerging properties in self-supervised vision transformers.
\newblock In \emph{2021 IEEE/CVF international conference on computer vision
  (ICCV)}, 9630--9640. IEEE.

\bibitem[{Chen et~al.(2025)Chen, Chen, Xu, Li, Dong, Sun, Jiang, Li, Yang, Zhao
  et~al.}]{chen2025dancetogether}
Chen, J.; Chen, M.; Xu, J.; Li, X.; Dong, J.; Sun, M.; Jiang, P.; Li, H.; Yang,
  Y.; Zhao, H.; et~al. 2025.
\newblock Dancetogether! identity-preserving multi-person interactive video
  generation.
\newblock \emph{arXiv preprint arXiv:2505.18078}.

\bibitem[{Chen et~al.(2024)Chen, Liu, He, Chen, Gan, Ma, Zhong, Zhang, Wang,
  Lin et~al.}]{chen2024mecd}
Chen, T.; Liu, H.; He, T.; Chen, Y.; Gan, C.; Ma, X.; Zhong, C.; Zhang, Y.;
  Wang, Y.; Lin, H.; et~al. 2024.
\newblock MECD: Unlocking multi-event causal discovery in video reasoning.
\newblock \emph{Advances in neural information processing systems}, 37:
  92554--92580.

\bibitem[{Cheng et~al.(2026)Cheng, Hou, Tao, and Liao}]{cheng2026video}
Cheng, J.; Hou, L.; Tao, X.; and Liao, J. 2026.
\newblock Video-as-answer: Predict and generate next video event with
  joint-grpo.
\newblock In \emph{Proceedings of the IEEE/CVF Conference on Computer Vision
  and Pattern Recognition}, 38915--38925.

\bibitem[{Fei et~al.(2024)Fei, Wu, Ji, Zhang, Zhang, Lee, and
  Hsu}]{fei2024video}
Fei, H.; Wu, S.; Ji, W.; Zhang, H.; Zhang, M.; Lee, M.-L.; and Hsu, W. 2024.
\newblock Video-of-thought: Step-by-step video reasoning from perception to
  cognition.
\newblock \emph{arXiv preprint arXiv:2501.03230}.

\bibitem[{Feng et~al.(2026)Feng, Gong, Li, Guo, Wang, Peng, Wu, Zhang, Wang,
  and Yue}]{feng2026video}
Feng, K.; Gong, K.; Li, B.; Guo, Z.; Wang, Y.; Peng, T.; Wu, J.; Zhang, X.;
  Wang, B.; and Yue, X. 2026.
\newblock Video-r1: Reinforcing video reasoning in mllms.
\newblock \emph{Advances in Neural Information Processing Systems}, 38:
  99114--99137.

\bibitem[{Guo et~al.(2025)Guo, Yang, Yang, Ma, Lin, Yang, Lin, and
  Jiang}]{guo2025long}
Guo, Y.; Yang, C.; Yang, Z.; Ma, Z.; Lin, Z.; Yang, Z.; Lin, D.; and Jiang, L.
  2025.
\newblock Long context tuning for video generation.
\newblock In \emph{Proceedings of the IEEE/CVF International Conference on
  Computer Vision}, 17281--17291.

\bibitem[{Han et~al.(2026)Han, Zhai, Yang, Cao, and Zha}]{han2026touch}
Han, G.; Zhai, W.; Yang, Y.; Cao, Y.; and Zha, Z.-J. 2026.
\newblock Touch: Text-guided controllable generation of free-form hand-object
  interactions.
\newblock In \emph{International Conference on Learning Representations},
  volume 2026, 79185--79210.

\bibitem[{He et~al.(2025)He, Liu, Li, Huang, Yu, Ouyang, and
  Liu}]{he2025cut2next}
He, J.; Liu, H.; Li, J.; Huang, Z.; Yu, Q.; Ouyang, W.; and Liu, Z. 2025.
\newblock Cut2next: Generating next shot via in-context tuning.
\newblock In \emph{Proceedings of the SIGGRAPH Asia 2025 Conference Papers},
  1--11.

\bibitem[{Hendrycks and Gimpel(2016)}]{hendrycks2016gaussian}
Hendrycks, D.; and Gimpel, K. 2016.
\newblock Gaussian error linear units (gelus).
\newblock \emph{arXiv preprint arXiv:1606.08415}.

\bibitem[{Huang et~al.(2024)Huang, He, Yu, Zhang, Si, Jiang, Zhang, Wu, Jin,
  Chanpaisit et~al.}]{huang2024vbench}
Huang, Z.; He, Y.; Yu, J.; Zhang, F.; Si, C.; Jiang, Y.; Zhang, Y.; Wu, T.;
  Jin, Q.; Chanpaisit, N.; et~al. 2024.
\newblock Vbench: Comprehensive benchmark suite for video generative models.
\newblock In \emph{2024 IEEE/CVF Conference on Computer Vision and Pattern
  Recognition (CVPR)}, 21807--21818. IEEE.

\bibitem[{Huang et~al.(2025)Huang, Yu, Chen, Qiu, Debevec, and
  Liu}]{huang2025vchain}
Huang, Z.; Yu, N.; Chen, G.; Qiu, H.; Debevec, P.; and Liu, Z. 2025.
\newblock Vchain: Chain-of-visual-thought for reasoning in video generation.
\newblock \emph{arXiv preprint arXiv:2510.05094}.

\bibitem[{Jiang et~al.(2024)Jiang, Wu, Yang, Si, Lin, Qiao, Loy, and
  Liu}]{jiang2024videobooth}
Jiang, Y.; Wu, T.; Yang, S.; Si, C.; Lin, D.; Qiao, Y.; Loy, C.~C.; and Liu, Z.
  2024.
\newblock Videobooth: Diffusion-based video generation with image prompts.
\newblock In \emph{2024 IEEE/CVF Conference on Computer Vision and Pattern
  Recognition (CVPR)}, 6689--6700. IEEE.

\bibitem[{Kingma and Ba(2014)}]{kingma2014adam}
Kingma, D.~P.; and Ba, J. 2014.
\newblock Adam: A method for stochastic optimization.
\newblock \emph{arXiv preprint arXiv:1412.6980}.

\bibitem[{Kong et~al.(2024)Kong, Tian, Zhang, Min, Dai, Zhou, Xiong, Li, Wu,
  Zhang et~al.}]{kong2024hunyuanvideo}
Kong, W.; Tian, Q.; Zhang, Z.; Min, R.; Dai, Z.; Zhou, J.; Xiong, J.; Li, X.;
  Wu, B.; Zhang, J.; et~al. 2024.
\newblock Hunyuanvideo: A systematic framework for large video generative
  models.
\newblock \emph{arXiv preprint arXiv:2412.03603}.

\bibitem[{Lei et~al.(2020)Lei, Yu, Berg, and Bansal}]{lei2020more}
Lei, J.; Yu, L.; Berg, T.; and Bansal, M. 2020.
\newblock What is more likely to happen next? video-and-language future event
  prediction.
\newblock In \emph{Proceedings of the 2020 conference on empirical methods in
  natural language processing (EMNLP)}, 8769--8784.

\bibitem[{Li et~al.(2024)Li, Li, Yang, Cao, Zhu, Cheng, and
  Chen}]{li2024dispose}
Li, H.; Li, Y.; Yang, Y.; Cao, J.; Zhu, Z.; Cheng, X.; and Chen, L. 2024.
\newblock Dispose: Disentangling pose guidance for controllable human image
  animation.
\newblock \emph{arXiv preprint arXiv:2412.09349}.

\bibitem[{Li et~al.(2023{\natexlab{a}})Li, Wei, Han, and Fan}]{li2023intentqa}
Li, J.; Wei, P.; Han, W.; and Fan, L. 2023{\natexlab{a}}.
\newblock Intentqa: Context-aware video intent reasoning.
\newblock In \emph{2023 IEEE/CVF International Conference on Computer Vision
  (ICCV)}, 11929--11940. IEEE.

\bibitem[{Li et~al.(2026)Li, Chen, Zhao, Schiffers, Liao, and
  Bhat}]{li2026happens}
Li, X.; Chen, Z.; Zhao, R.; Schiffers, F.; Liao, Z.; and Bhat, V. 2026.
\newblock What happens next? next scene prediction with a unified video model.
\newblock In \emph{Proceedings of the IEEE/CVF Conference on Computer Vision
  and Pattern Recognition}, 5693--5703.

\bibitem[{Li et~al.(2023{\natexlab{b}})Li, Zhu, Han, Hou, Guo, and
  Cheng}]{li2023amt}
Li, Z.; Zhu, Z.-L.; Han, L.-H.; Hou, Q.; Guo, C.-L.; and Cheng, M.-M.
  2023{\natexlab{b}}.
\newblock Amt: All-pairs multi-field transforms for efficient frame
  interpolation.
\newblock In \emph{2023 IEEE/CVF Conference on Computer Vision and Pattern
  Recognition (CVPR)}, 9801--9810. IEEE.

\bibitem[{Liang et~al.(2025)Liang, Su, Zhu, Liang, and
  Tong}]{liang2025videvent}
Liang, B.; Su, Q.; Zhu, S.; Liang, Y.; and Tong, C. 2025.
\newblock VidEvent: A Large Dataset for Understanding Dynamic Evolution of
  Events in Videos.
\newblock \emph{Proceedings of the AAAI Conference on Artificial Intelligence},
  39(5): 5128--5136.

\bibitem[{Liang et~al.(2024)Liang, Jiang, Wang, Pan, Chen, Chu, Liu, Fu, Wang,
  and Qin}]{liang2024guide}
Liang, J.; Jiang, S.; Wang, Z.; Pan, H.; Chen, Z.; Chu, Z.; Liu, M.; Fu, R.;
  Wang, Z.; and Qin, B. 2024.
\newblock GUIDE: a guideline-guided dataset for instructional video
  comprehension.
\newblock \emph{arXiv preprint arXiv:2406.18227}.

\bibitem[{Lin et~al.(2023)Lin, Zala, Cho, and Bansal}]{lin2023videodirectorgpt}
Lin, H.; Zala, A.; Cho, J.; and Bansal, M. 2023.
\newblock Videodirectorgpt: Consistent multi-scene video generation via
  llm-guided planning.
\newblock \emph{arXiv preprint arXiv:2309.15091}.

\bibitem[{Lipman et~al.(2022)Lipman, Chen, Ben-Hamu, Nickel, and
  Le}]{lipman2022flow}
Lipman, Y.; Chen, R.~T.; Ben-Hamu, H.; Nickel, M.; and Le, M. 2022.
\newblock Flow matching for generative modeling.
\newblock \emph{arXiv preprint arXiv:2210.02747}.

\bibitem[{Liu et~al.(2025)Liu, Ma, Li, Chen, Liu, Li, Zhou, He, and
  Wu}]{liu2025phantom}
Liu, L.; Ma, T.; Li, B.; Chen, Z.; Liu, J.; Li, G.; Zhou, S.; He, Q.; and Wu,
  X. 2025.
\newblock Phantom: Subject-consistent video generation via cross-modal
  alignment.
\newblock In \emph{2025 IEEE/CVF International Conference on Computer Vision
  (ICCV)}, 14951--14961. IEEE.

\bibitem[{Liu, Gong, and Liu(2022)}]{liu2022flow}
Liu, X.; Gong, C.; and Liu, Q. 2022.
\newblock Flow straight and fast: Learning to generate and transfer data with
  rectified flow.
\newblock \emph{arXiv preprint arXiv:2209.03003}.

\bibitem[{Long et~al.(2024)Long, Qiu, Yao, and Mei}]{long2024videostudio}
Long, F.; Qiu, Z.; Yao, T.; and Mei, T. 2024.
\newblock Videostudio: Generating consistent-content and multi-scene videos.
\newblock In \emph{European Conference on Computer Vision}, 468--485. Springer.

\bibitem[{Luo et~al.(2025)Luo, Lin, Zhang, Wu, Fang, Chen, and
  Tang}]{luo2025univid}
Luo, J.; Lin, J.; Zhang, Z.; Wu, B.; Fang, M.; Chen, L.; and Tang, H. 2025.
\newblock Univid: The open-source unified video model.
\newblock \emph{arXiv preprint arXiv:2509.24200}.

\bibitem[{Luo et~al.(2026)Luo, Shi, Zhuang, Chen, Liu, Wang, Wan, and
  Xue}]{luo2026shotstream}
Luo, Y.; Shi, X.; Zhuang, J.; Chen, Y.; Liu, Q.; Wang, X.; Wan, P.; and Xue, T.
  2026.
\newblock Shotstream: Streaming multi-shot video generation for interactive
  storytelling.
\newblock \emph{arXiv preprint arXiv:2603.25746}.

\bibitem[{Ma et~al.(2025)Ma, Wu, Sun, and Li}]{ma2025hpsv3}
Ma, Y.; Wu, X.; Sun, K.; and Li, H. 2025.
\newblock Hpsv3: Towards wide-spectrum human preference score.
\newblock In \emph{2025 IEEE/CVF International Conference on Computer Vision
  (ICCV)}, 15086--15095. IEEE.

\bibitem[{Ma et~al.(2024)Ma, Zhou, Wang, Yeh, Li, Yang, Dong, Keutzer, and
  Feng}]{ma2024magic}
Ma, Z.; Zhou, D.; Wang, X.-S.; Yeh, C.-H.; Li, X.; Yang, H.; Dong, Z.; Keutzer,
  K.; and Feng, J. 2024.
\newblock Magic-me: Identity-specific video customized diffusion.
\newblock In \emph{European Conference on Computer Vision}, 19--37. Springer.

\bibitem[{Meng et~al.(2026)Meng, Ouyang, Yu, Wang, Wang, Cheng, Wang, Ma, Li,
  Chen et~al.}]{meng2026holocine}
Meng, Y.; Ouyang, H.; Yu, Y.; Wang, Q.; Wang, W.; Cheng, K.~L.; Wang, H.; Ma,
  S.; Li, Y.; Chen, C.; et~al. 2026.
\newblock Holocine: Holistic generation of cinematic multi-shot long video
  narratives.
\newblock In \emph{Proceedings of the IEEE/CVF conference on computer vision
  and pattern recognition}, 461--471.

\bibitem[{Min et~al.(2024)Min, Buch, Nagrani, Cho, and Schmid}]{min2024morevqa}
Min, J.; Buch, S.; Nagrani, A.; Cho, M.; and Schmid, C. 2024.
\newblock Morevqa: Exploring modular reasoning models for video question
  answering.
\newblock In \emph{2024 IEEE/CVF Conference on Computer Vision and Pattern
  Recognition (CVPR)}, 13235--13245. IEEE.

\bibitem[{Peebles and Xie(2023)}]{peebles2023scalable}
Peebles, W.; and Xie, S. 2023.
\newblock Scalable diffusion models with transformers.
\newblock In \emph{2023 IEEE/CVF International Conference on Computer Vision
  (ICCV)}, 4172--4182. IEEE.

\bibitem[{Polyak et~al.(2024)Polyak, Zohar, Brown, Tjandra, Sinha, Lee, Vyas,
  Shi, Ma, Chuang et~al.}]{polyak2024movie}
Polyak, A.; Zohar, A.; Brown, A.; Tjandra, A.; Sinha, A.; Lee, A.; Vyas, A.;
  Shi, B.; Ma, C.-Y.; Chuang, C.-Y.; et~al. 2024.
\newblock Movie gen: A cast of media foundation models.
\newblock \emph{arXiv preprint arXiv:2410.13720}.

\bibitem[{Radford et~al.(2021)Radford, Kim, Hallacy, Ramesh, Goh, Agarwal,
  Sastry, Askell, Mishkin, Clark et~al.}]{radford2021learning}
Radford, A.; Kim, J.~W.; Hallacy, C.; Ramesh, A.; Goh, G.; Agarwal, S.; Sastry,
  G.; Askell, A.; Mishkin, P.; Clark, J.; et~al. 2021.
\newblock Learning transferable visual models from natural language
  supervision.
\newblock In \emph{International conference on machine learning}, 8748--8763.
  PmLR.

\bibitem[{Shao et~al.(2025)Shao, Zhai, Yang, Luo, Cao, and Zha}]{shao2025great}
Shao, Y.; Zhai, W.; Yang, Y.; Luo, H.; Cao, Y.; and Zha, Z.-J. 2025.
\newblock Great: Geometry-intention collaborative inference for open-vocabulary
  3d object affordance grounding.
\newblock In \emph{2025 IEEE/CVF Conference on Computer Vision and Pattern
  Recognition (CVPR)}, 17326--17336. IEEE.

\bibitem[{Shen et~al.(2025)Shen, Maksutova, Li, and
  Unberath}]{shen2025counterfactual}
Shen, Y.; Maksutova, A.; Li, C.; and Unberath, M. 2025.
\newblock Counterfactual world models via digital twin-conditioned video
  diffusion.
\newblock \emph{arXiv preprint arXiv:2511.17481}.

\bibitem[{Spyrou et~al.(2025)Spyrou, Vlontzos, Pegios, Melistas, Gkouti,
  Panagakis, Papanastasiou, and Tsaftaris}]{spyrou2025causally}
Spyrou, N.; Vlontzos, A.; Pegios, P.; Melistas, T.; Gkouti, N.; Panagakis, Y.;
  Papanastasiou, G.; and Tsaftaris, S.~A. 2025.
\newblock Causally steered diffusion for automated video counterfactual
  generation.
\newblock \emph{arXiv preprint arXiv:2506.14404}.

\bibitem[{Tan et~al.(2025)Tan, Yang, Qin, Gong, Yang, and Li}]{tan2025omni}
Tan, Z.; Yang, H.; Qin, L.; Gong, J.; Yang, M.; and Li, H. 2025.
\newblock Omni-video: Democratizing unified video understanding and generation.
\newblock \emph{arXiv preprint arXiv:2507.06119}.

\bibitem[{Teng et~al.(2025)Teng, Jia, Sun, Li, Li, Tang, Han, Zhang, Zhang, Luo
  et~al.}]{teng2025magi}
Teng, H.; Jia, H.; Sun, L.; Li, L.; Li, M.; Tang, M.; Han, S.; Zhang, T.;
  Zhang, W.; Luo, W.; et~al. 2025.
\newblock Magi-1: Autoregressive video generation at scale.
\newblock \emph{arXiv preprint arXiv:2505.13211}.

\bibitem[{Wan et~al.(2025)Wan, Wang, Ai, Wen, Mao, Xie, Chen, Yu, Zhao, Yang
  et~al.}]{wan2025wan}
Wan, T.; Wang, A.; Ai, B.; Wen, B.; Mao, C.; Xie, C.-W.; Chen, D.; Yu, F.;
  Zhao, H.; Yang, J.; et~al. 2025.
\newblock Wan: Open and advanced large-scale video generative models.
\newblock \emph{arXiv preprint arXiv:2503.20314}.

\bibitem[{Wang et~al.(2026)Wang, Liu, Liu, Du, Kawaguchi, Wang, and
  Pang}]{wang2026fostering}
Wang, H.; Liu, H.; Liu, X.; Du, C.; Kawaguchi, K.; Wang, Y.; and Pang, T. 2026.
\newblock Fostering video reasoning via next-event prediction.
\newblock In \emph{International Conference on Learning Representations},
  volume 2026, 31524--31570.

\bibitem[{Wei et~al.(2026)Wei, Liu, Ye, Wang, Wang, Wan, Gai, and
  Chen}]{wei2026univideo}
Wei, C.; Liu, Q.; Ye, Z.; Wang, Q.; Wang, X.; Wan, P.; Gai, K.; and Chen, W.
  2026.
\newblock Univideo: Unified understanding, generation, and editing for videos.
\newblock In \emph{International Conference on Learning Representations},
  volume 2026, 113905--113933.

\bibitem[{Wu et~al.(2024)Wu, Yu, Chen, Tenenbaum, and Gan}]{wu2024star}
Wu, B.; Yu, S.; Chen, Z.; Tenenbaum, J.~B.; and Gan, C. 2024.
\newblock Star: A benchmark for situated reasoning in real-world videos.
\newblock \emph{arXiv preprint arXiv:2405.09711}.

\bibitem[{Wu, Zhu, and Shou(2025)}]{wu2025automated}
Wu, W.; Zhu, Z.; and Shou, M.~Z. 2025.
\newblock Automated movie generation via multi-agent cot planning.
\newblock \emph{arXiv preprint arXiv:2503.07314}.

\bibitem[{Xiao et~al.(2025)Xiao, Cheng, Qi, Gui, Zhao, Lin, Cen, Ma, Yuille,
  and Jiang}]{xiao2025videoauteur}
Xiao, J.; Cheng, F.; Qi, L.; Gui, L.; Zhao, Y.; Lin, S.; Cen, J.; Ma, Z.;
  Yuille, A.; and Jiang, L. 2025.
\newblock Videoauteur: Towards long narrative video generation.
\newblock In \emph{2025 IEEE/CVF International Conference on Computer Vision
  (ICCV)}, 19163--19173. IEEE.

\bibitem[{Xiao et~al.(2021)Xiao, Shang, Yao, and Chua}]{xiao2021next}
Xiao, J.; Shang, X.; Yao, A.; and Chua, T.-S. 2021.
\newblock Next-qa: Next phase of question-answering to explaining temporal
  actions.
\newblock In \emph{2021 IEEE/CVF Conference on Computer Vision and Pattern
  Recognition (CVPR)}, 9772--9781. IEEE.

\bibitem[{Yang et~al.(2025{\natexlab{a}})Yang, Fan, Sun, Li, Zeng, Han, Zhai,
  Liu, Cao, and Zha}]{yang2025videogen}
Yang, Y.; Fan, K.; Sun, S.; Li, H.; Zeng, A.; Han, F.; Zhai, W.; Liu, W.; Cao,
  Y.; and Zha, Z.-J. 2025{\natexlab{a}}.
\newblock Videogen-eval: Agent-based system for video generation evaluation.
\newblock \emph{arXiv preprint arXiv:2503.23452}.

\bibitem[{Yang et~al.(2025{\natexlab{b}})Yang, Liu, Lu, Zhao, Wu, Zhai, Yi,
  Cao, Ma, Zha et~al.}]{yang2025sigman}
Yang, Y.; Liu, F.; Lu, Y.; Zhao, Q.; Wu, P.; Zhai, W.; Yi, R.; Cao, Y.; Ma, L.;
  Zha, Z.-J.; et~al. 2025{\natexlab{b}}.
\newblock Sigman: Scaling 3d human gaussian generation with millions of assets.
\newblock In \emph{2025 IEEE/CVF International Conference on Computer Vision
  (ICCV)}, 5122--5133. IEEE.

\bibitem[{Yang et~al.(2023)Yang, Zhai, Luo, Cao, Luo, and
  Zha}]{yang2023grounding}
Yang, Y.; Zhai, W.; Luo, H.; Cao, Y.; Luo, J.; and Zha, Z.-J. 2023.
\newblock Grounding 3d object affordance from 2d interactions in images.
\newblock In \emph{Proceedings of the IEEE/CVF International Conference on
  Computer Vision}, 10905--10915.

\bibitem[{Yang et~al.(2024{\natexlab{a}})Yang, Zhai, Luo, Cao, and
  Zha}]{yang2024lemon}
Yang, Y.; Zhai, W.; Luo, H.; Cao, Y.; and Zha, Z.-J. 2024{\natexlab{a}}.
\newblock Lemon: Learning 3d human-object interaction relation from 2d images.
\newblock In \emph{Proceedings of the IEEE/CVF Conference on Computer Vision
  and Pattern Recognition}, 16284--16295.

\bibitem[{Yang et~al.(2024{\natexlab{b}})Yang, Zhai, Wang, Yu, Cao, and
  Zha}]{yang2024egochoir}
Yang, Y.; Zhai, W.; Wang, C.; Yu, C.; Cao, Y.; and Zha, Z.-J.
  2024{\natexlab{b}}.
\newblock Egochoir: Capturing 3d human-object interaction regions from
  egocentric views.
\newblock \emph{Advances in Neural Information Processing Systems}, 37:
  54529--54557.

\bibitem[{Yang et~al.(2026)Yang, Zhang, Pi, Zeng, Guo, Xu, Zhai, Cao, and
  Zha}]{yang2026gloria}
Yang, Y.; Zhang, F.; Pi, H.; Zeng, A.; Guo, S.; Xu, G.; Zhai, W.; Cao, Y.; and
  Zha, Z.-J. 2026.
\newblock Gloria: Consistent Character Video Generation via Content Anchors.
\newblock In \emph{Proceedings of the IEEE/CVF Conference on Computer Vision
  and Pattern Recognition}, 36724--36735.

\bibitem[{Yang et~al.(2025{\natexlab{c}})Yang, Teng, Zheng, Ding, Huang, Xu,
  Yang, Hong, Zhang, Feng et~al.}]{yang2025cogvideox}
Yang, Z.; Teng, J.; Zheng, W.; Ding, M.; Huang, S.; Xu, J.; Yang, Y.; Hong, W.;
  Zhang, X.; Feng, G.; et~al. 2025{\natexlab{c}}.
\newblock Cogvideox: Text-to-video diffusion models with an expert transformer.
\newblock In \emph{International Conference on Learning Representations},
  volume 2025, 83048--83077.

\bibitem[{Yi et~al.(2019)Yi, Gan, Li, Kohli, Wu, Torralba, and
  Tenenbaum}]{yi2019clevrer}
Yi, K.; Gan, C.; Li, Y.; Kohli, P.; Wu, J.; Torralba, A.; and Tenenbaum, J.~B.
  2019.
\newblock Clevrer: Collision events for video representation and reasoning.
\newblock \emph{arXiv preprint arXiv:1910.01442}.

\bibitem[{Yu et~al.(2025)Yu, Zhai, Yang, Cao, and Zha}]{yu2025hero}
Yu, C.; Zhai, W.; Yang, Y.; Cao, Y.; and Zha, Z.-J. 2025.
\newblock Hero: Human reaction generation from videos.
\newblock In \emph{2025 IEEE/CVF International Conference on Computer Vision
  (ICCV)}, 10262--10274. IEEE.

\bibitem[{Yuan et~al.(2025)Yuan, Huang, He, Ge, Shi, Chen, Luo, and
  Yuan}]{yuan2025identity}
Yuan, S.; Huang, J.; He, X.; Ge, Y.; Shi, Y.; Chen, L.; Luo, J.; and Yuan, L.
  2025.
\newblock Identity-preserving text-to-video generation by frequency
  decomposition.
\newblock In \emph{2025 IEEE/CVF Conference on Computer Vision and Pattern
  Recognition (CVPR)}, 12978--12988. IEEE.

\bibitem[{Zeng et~al.(2026)Zeng, Yang, Ge, Zhang, Xu, Lin, Gu, Pi, Li, Shi
  et~al.}]{zeng2026lpm}
Zeng, A.; Yang, C.; Ge, C.; Zhang, E.; Xu, G.; Lin, G.; Gu, G.; Pi, J.; Li, L.;
  Shi, M.; et~al. 2026.
\newblock Lpm 1.0: Video-based character performance model.
\newblock \emph{arXiv preprint arXiv:2604.07823}.

\bibitem[{Zeng et~al.(2024)Zeng, Yang, Chen, and Liu}]{zeng2024dawn}
Zeng, A.; Yang, Y.; Chen, W.; and Liu, W. 2024.
\newblock The dawn of video generation: Preliminary explorations with sora-like
  models.
\newblock \emph{arXiv preprint arXiv:2410.05227}.

\bibitem[{Zhang et~al.(2025)Zhang, Jiang, Wang, Fang, Zhi, Yan, Kang, Lu, and
  Pan}]{zhang2025storymem}
Zhang, K.; Jiang, L.; Wang, A.; Fang, J.~Z.; Zhi, T.; Yan, Q.; Kang, H.; Lu,
  X.; and Pan, X. 2025.
\newblock Storymem: Multi-shot long video storytelling with memory.
\newblock \emph{arXiv preprint arXiv:2512.19539}.

\bibitem[{Zhang et~al.(2026)Zhang, Jia, Liu, Weng, Li, and
  Shi}]{zhang2026stage}
Zhang, P.; Jia, Z.; Liu, K.; Weng, S.; Li, S.; and Shi, B. 2026.
\newblock Stage: Storyboard-anchored generation for cinematic multi-shot
  narrative.
\newblock In \emph{Proceedings of the IEEE/CVF Conference on Computer Vision
  and Pattern Recognition}, 659--669.

\bibitem[{Zheng et~al.(2024)Zheng, Xu, Huang, Ma, Liu, Shu, Pang, Tang, Chen,
  Yang et~al.}]{zheng2024videogen}
Zheng, M.; Xu, Y.; Huang, H.; Ma, X.; Liu, Y.; Shu, W.; Pang, Y.; Tang, F.;
  Chen, Q.; Yang, H.; et~al. 2024.
\newblock Videogen-of-thought: Step-by-step generating multi-shot video with
  minimal manual intervention.
\newblock \emph{arXiv preprint arXiv:2412.02259}.

\bibitem[{Zhuang et~al.(2024)Zhuang, Li, Chen, Wang, Liu, Qiao, and
  Wang}]{zhuang2024vlogger}
Zhuang, S.; Li, K.; Chen, X.; Wang, Y.; Liu, Z.; Qiao, Y.; and Wang, Y. 2024.
\newblock Vlogger: Make your dream a vlog.
\newblock In \emph{2024 IEEE/CVF Conference on Computer Vision and Pattern
  Recognition (CVPR)}, 8806--8817. IEEE.

\end{thebibliography}

\clearpage
\appendix

\section{Data Construction, Composition, and Quality}
\label{sec:supp_data_quality}

This section presents the prompt-verification protocol, the composition
of the resulting dataset, and the human audits and judge validation used
to assess data quality.

\subsection{Prompt-Verification Protocol}
\label{sec:supp_data_construction}

The prompt-verification judge evaluates whether the predicted target-event
description matches the event depicted in the ground-truth target video.
The judge further verifies that the target event can be accurately inferred
only by jointly considering the context video, the prompt instruction,
and the starting frame---penalizing cases in which the event can be deduced
from any subset of these inputs. This evaluation uses the following
five-point rubric:
\begin{itemize}
  \item[$5$] Semantically identical target event and outcome, with the
        inference strictly relying on all three inputs jointly.
  \item[$4$] Same target event with minor differences in detail, provided
        the inference still requires all three inputs jointly.
  \item[$3$] Partially related event with significant differences, or
        the inference relies too heavily on a single input (e.g., the
        prompt instruction alone reveals the action).
  \item[$2$] Shares the broad theme but predicts a clearly incorrect event,
        or the target event is completely deducible without considering
        all three inputs jointly.
  \item[$1$] Completely unrelated event or outcome.
\end{itemize}
A pair passes prompt verification only if it receives a score of at
least $4$; otherwise, it undergoes pair regeneration to resolve
ambiguities or remove overly explicit cues before being re-evaluated.
Pairs that remain below the threshold after regeneration and
re-evaluation are discarded, whereas accepted pairs are assigned a
quality score by the judge.

\subsection{Dataset Composition}
\label{sec:supp_dataset_composition}

The final dataset contains 110K samples. Each sample includes three
generation inputs---a context video, a prompt instruction, and a
starting frame---together with a target video used as supervision.
We characterize the dataset composition using two hierarchical taxonomies:
task categories and cross-shot relations. In the sunburst charts in
Fig.~\ref{fig:data_pipeline}, the inner rings represent the primary task
categories and relation types, while the outer rings represent their
fine-grained subcategories and subtypes.
Tables~\ref{tab:supp_task_distribution} and
\ref{tab:supp_relation_distribution} report the exact counts and
percentages corresponding to all sectors. The former lists task categories
and subcategories, while the latter lists cross-shot relation types and
subtypes. Bold rows report primary-category or relation-type totals, and
indented rows report their fine-grained subdivisions. Percentages are
computed over the full dataset and rounded independently.

The five primary relation types characterize how the context and target
videos are logically connected. \emph{Progression} covers transitions
between stages, steps, locations, or tools in a broader
process. \emph{Parallel} describes analogous or alternative actions or
events. \emph{Causal} captures cause--effect links, state changes, and
action--result relations. \emph{Conditional} associates the target action
with a prerequisite or trigger established in the context.
\emph{Overview/Detail} connects an overall scene or process with one of
its components or details.

\begin{table}[t]
  \centering
  \setlength{\tabcolsep}{1pt}
  \renewcommand{\arraystretch}{0.93}
  \footnotesize
  \begin{tabular}{@{}lrr@{}}
    \toprule
    Task category / subcategory & \# Samples & \% of dataset \\
    \midrule
    \textbf{Cooking \& Beverages} & \textbf{26{,}009} & \textbf{23.02} \\
    \quad Recipes \& Tutorials & 10{,}323 & 9.14 \\
    \quad Savory Mains & 5{,}398 & 4.78 \\
    \quad Drinks, Coffee \& Cocktails & 4{,}604 & 4.07 \\
    \quad Desserts \& Sweets & 3{,}498 & 3.10 \\
    \quad Prep, Pickling \& Sides & 2{,}186 & 1.93 \\
    \addlinespace[1.5pt]
    \textbf{Device Setup \& Operation} & \textbf{11{,}196} & \textbf{9.91} \\
    \quad Office \& Consumer Electronics & 6{,}119 & 5.42 \\
    \quad Imaging Equipment & 5{,}077 & 4.49 \\
    \addlinespace[1.5pt]
    \textbf{Furniture \& Large Assembly} & \textbf{26{,}086} & \textbf{23.09} \\
    \quad Bedside \& Lounge Furniture & 9{,}572 & 8.47 \\
    \quad Storage, Shelving \& Cages & 8{,}882 & 7.86 \\
    \quad Tables, Chairs \& Small Furniture & 7{,}632 & 6.75 \\
    \addlinespace[1.5pt]
    \textbf{Household, Crafts \& Personal Care} & \textbf{6{,}244} & \textbf{5.53} \\
    \quad Cleaning \& Household & 2{,}354 & 2.08 \\
    \quad Crafts \& DIY & 2{,}119 & 1.88 \\
    \quad Personal Care \& Styling & 1{,}771 & 1.57 \\
    \addlinespace[1.5pt]
    \textbf{Mechanical, Parts \& Model Assembly} & \textbf{14{,}720} & \textbf{13.03} \\
    \quad Mechanical \& Model Assembly & 12{,}558 & 11.11 \\
    \quad Parts \& Structural Assembly & 2{,}162 & 1.91 \\
    \addlinespace[1.5pt]
    \textbf{Medical, Lab \& Safety} & \textbf{1{,}344} & \textbf{1.19} \\
    \quad Medical Care & 762 & 0.67 \\
    \quad Lab Work \& Instruments & 463 & 0.41 \\
    \quad Safety Procedures & 119 & 0.11 \\
    \addlinespace[1.5pt]
    \textbf{Narrative \& Scene Progression} & \textbf{16{,}238} & \textbf{14.37} \\
    \quad Film \& Event Progression & 12{,}316 & 10.90 \\
    \quad General Scene Progression & 3{,}922 & 3.47 \\
    \addlinespace[1.5pt]
    \textbf{Repair \& Replacement} & \textbf{8{,}280} & \textbf{7.33} \\
    \quad Home \& General Maintenance & 4{,}407 & 3.90 \\
    \quad Automotive \& Mechanical Maintenance & 2{,}808 & 2.49 \\
    \quad Electronics \& Device Repair & 1{,}065 & 0.94 \\
    \addlinespace[1.5pt]
    \textbf{Sports, Outdoors \& Gardening} & \textbf{2{,}866} & \textbf{2.54} \\
    \quad Sports Skills \& Activities & 1{,}619 & 1.43 \\
    \quad Gardening \& Planting & 844 & 0.75 \\
    \quad Outdoors, Survival \& Camping & 403 & 0.36 \\
    \bottomrule
  \end{tabular}
  \caption{Fine-grained task-category distribution in the dataset.}
  \label{tab:supp_task_distribution}
\end{table}

\begin{table}[t]
  \centering
  \setlength{\tabcolsep}{3pt}
  \renewcommand{\arraystretch}{0.93}
  \footnotesize
  \begin{tabular}{@{}lrr@{}}
    \toprule
    Relation type / subtype & \# Samples & \% of dataset \\
    \midrule
    \textbf{Progression} & \textbf{32{,}444} & \textbf{28.72} \\
    \quad Spatial Transition & 11{,}912 & 10.54 \\
    \quad Object/Tool Shift & 8{,}611 & 7.62 \\
    \quad Process Continuation & 8{,}388 & 7.42 \\
    \quad Step Continuation & 2{,}487 & 2.20 \\
    \quad Post-Completion Step & 1{,}046 & 0.93 \\
    \addlinespace[1.5pt]
    \textbf{Parallel} & \textbf{24{,}128} & \textbf{21.36} \\
    \quad Action Analogy & 18{,}720 & 16.57 \\
    \quad Object Substitution & 2{,}845 & 2.52 \\
    \quad Event Restatement & 18 & 0.02 \\
    \quad Action Variant & 2{,}545 & 2.25 \\
    \addlinespace[1.5pt]
    \textbf{Causal} & \textbf{22{,}605} & \textbf{20.01} \\
    \quad Preparation-to-Action & 15{,}794 & 13.98 \\
    \quad Cause-to-Effect & 1{,}842 & 1.63 \\
    \quad State Change & 3{,}337 & 2.95 \\
    \quad Action-to-Result & 1{,}632 & 1.44 \\
    \addlinespace[1.5pt]
    \textbf{Conditional} & \textbf{18{,}027} & \textbf{15.96} \\
    \quad Preparation Condition & 11{,}401 & 10.09 \\
    \quad Post-Action Inspection & 4{,}326 & 3.83 \\
    \quad Conditional Response & 138 & 0.12 \\
    \quad Trigger-to-Action & 2{,}162 & 1.91 \\
    \addlinespace[1.5pt]
    \textbf{Overview/Detail} & \textbf{15{,}779} & \textbf{13.97} \\
    \quad Whole-Part & 85 & 0.08 \\
    \quad Component Expansion & 4{,}155 & 3.68 \\
    \quad Component Parallelism & 3{,}279 & 2.90 \\
    \quad Component Transition & 6{,}337 & 5.61 \\
    \quad Whole-to-Part Focus & 793 & 0.70 \\
    \quad Whole-to-Part Progression & 1{,}130 & 1.00 \\
    \bottomrule
  \end{tabular}
  \caption{Fine-grained cross-shot relation distribution in the dataset.}
  \label{tab:supp_relation_distribution}
\end{table}

\subsection{Data-Pipeline Human Audits and Judge Validation}
\label{sec:supp_data_audits}

We assess the data pipeline through two independent human audits and a
separate judge-validation study: a post-pair-generation audit of
candidate pairs, a post-prompt-verification audit of the final dataset,
and a validation study comparing the judge with a three-annotator
consensus. This section
details the sampling designs, rubrics, and resulting statistics for
each evaluation.

\paragraph{Post-Pair-Generation Audit.}
A uniform random sample of $1{,}500$ candidate pairs, drawn between
Prompt Generation and Prompt Verification, is audited by five
annotators. Each annotator labels a disjoint set of $250$ pairs; an
additional set of $250$ pairs is
independently labeled by three of the five annotators to estimate
Fleiss' $\kappa$. We apply two binary rubrics: \emph{video quality}
(valid shot boundaries and the absence of compression artifacts, HDR
encoding, incomplete clips, or text and logos that occlude salient
regions) and \emph{prompt validity} (the prompt instruction accurately describes
the cross-shot relation, and the target event is inferable only by jointly
considering the context video, the prompt instruction, and the starting frame). A pair
passes only when both rubrics
are satisfied. Per-rubric and composite pass rates, together with
Fleiss' $\kappa$ on the overlap subset, are reported in
Tab.~\ref{tab:supp_data_audits}.

\paragraph{Post-Prompt-Verification Audit.}
A uniform random sample of $1{,}000$ pairs from the final dataset is
audited by four annotators after prompt verification. Each annotator
labels a disjoint set of $200$ pairs; an additional set of $200$ pairs is
independently labeled by three of the four annotators to estimate
Fleiss' $\kappa$. We apply two binary rubrics: \emph{target-event inferability}
(the target event can be reliably inferred only by jointly considering
the context video, the prompt instruction, and the starting frame, not from any
subset of these inputs) and \emph{target-video faithfulness} (the target
video faithfully depicts the annotated target event). Per-rubric and
composite pass rates, together with Fleiss' $\kappa$, are reported in
Tab.~\ref{tab:supp_data_audits}.

\paragraph{Prompt-Verification Judge Validation.}
To validate the prompt-verification judge as an automatic quality gate,
we draw a stratified $500$-pair validation set: $250$ judge-accepted
and $250$ judge-rejected pairs, uniformly sampled within each
stratum. Three annotators independently label every pair using the
inferability rubric. Their majority vote defines the human consensus,
which is then compared with the judge's decision. Results are reported in
Tab.~\ref{tab:supp_data_audits}.

\begin{table*}[t]
  \centering
  \setlength{\tabcolsep}{1mm}
  \renewcommand{\arraystretch}{1.1}
  {\footnotesize
  \begin{tabular}{lcccccc}
    \toprule
    Stage & $n$ & Annotators & Rubric A & Rubric B
          & \begin{tabular}{@{}c@{}}Composite /\\Agreement\end{tabular} & $\kappa$ \\
    \midrule
    Post-pair generation
      & 1{,}500 & 5
      & 0.936 & 0.752
      & \textbf{0.716} & 0.71 \\
    Post-prompt verification
      & 1{,}000 & 4
      & 0.927 & 0.934
      & \textbf{0.893} & 0.74 \\
    Judge validation
      & \phantom{0}500 & 3
      & --- & ---
      & \textbf{0.916} & 0.78 \\
    \bottomrule
  \end{tabular}}
  \caption{Data-pipeline quality-gate results.
  For the post-pair-generation audit, Rubric A and Rubric B denote
  video quality and prompt validity; for the post-prompt-verification
  audit, they denote target-event inferability and target-video
  faithfulness. Composite is the rate of pairs passing both rubrics.
  For judge validation, Agreement is the rate of judge decisions
  matching human consensus. $\kappa$ denotes Fleiss'
  inter-annotator agreement for the two audits and Cohen's $\kappa$
  between the judge and human consensus for judge validation.}
  \label{tab:supp_data_audits}
\end{table*}

\section{Experimental Setup}
\label{sec:supp_setup}

This section outlines the model configuration, the baseline adaptations,
and the detailed metric definitions used throughout our experiments.

\subsection{Implementation Details}
\label{sec:supp_impl}

This section extends the backbone and generation specifications in
Sec.~\ref{sec:method} with details on the encoders, projection layers,
optimization, and data preprocessing.

\paragraph{Encoders and Projectors.} The CLIP image encoder $\Phi$~\cite{radford2021learning}
extracts $L_c{=}257$ tokens of dimension $d_c{=}1280$ from the
penultimate layer to prioritize semantic content over
classification-specific activations; a two-layer MLP projects these
tokens into the DiT conditioning space. The MCG projector maps the VLM
hidden states $\mathbf{H}$ into the dense multimodal cues
$P(\mathbf{H})$ using a two-layer MLP with a GELU non-linearity~\cite{hendrycks2016gaussian}.
VM encodes $K{=}4$ frames uniformly sampled from $V$ as VAE latent
slots and prepends them as a context-video prefix.

\paragraph{Frozen Components.} The video VAE $\mathcal{E}$, the DiT
text encoder $\tau$, the CLIP image encoder $\Phi$, and the VLM
$\Psi$ are kept frozen throughout training.

\paragraph{Optimization.} We use the Adam optimizer~\cite{kingma2014adam}
($\beta_1{=}0.9$, $\beta_2{=}0.999$, $\epsilon{=}10^{-8}$) with a
constant learning rate of $10^{-5}$, no warmup, a weight decay of
$5\times10^{-3}$, and a maximum gradient norm of $1.0$. The
rectified-flow noise level $\sigma$ is sampled from a shifted
flow-matching schedule~\cite{liu2022flow,lipman2022flow}
with a shift factor of $5$. For reproducibility,
we use a fixed random seed of $42$ throughout the experiments.

\paragraph{Input Preprocessing.} We apply cover-resize followed by center
cropping to match the training resolution. HDR-encoded clips are excluded
during data construction to maintain visual compatibility.

\paragraph{Compute and Sampling Configuration.} LogiShot uses the $14$B
Wan-family DiT backbone~\cite{wan2025wan}. Because $\Psi$, $\Phi$, $\mathcal{E}$, and
$\tau$ remain frozen, gradients are computed only for the DiT denoiser
and conditioning projectors. Training is conducted on $32$ H800 GPUs.
During inference, all methods produce $49$-frame
clips at $480{\times}832$ resolution and $16$ fps.

\subsection{Baseline Adaptation Details}
\label{sec:supp_baseline_adaptation}

Each baseline is adapted to use the same three inputs: the context video,
the prompt instruction, and a starting frame.
In all three adaptations, the VLM jointly processes these inputs when
producing the target-event description.

\paragraph{MLLM+I2V.}
This baseline serves as our caption-based reference: Qwen2.5-VL~\cite{bai2025qwen25vltechnicalreport}
generates a detailed target-event description from the three inputs;
a standard Wan2.1-14B image-to-video generator~\cite{wan2025wan}
then produces the target video conditioned on the resulting description
and the starting frame. For MLLM+I2V, context-video information reaches
the generator only through the decoded description, without direct access
to context-video features or latents. It generates $49$-frame clips at
$480{\times}832$ resolution and $16$~fps using $50$ denoising steps.

\paragraph{VANS~\cite{cheng2026video}.}
The adapted VANS uses the target-event description produced by its VLM
from the three inputs and VAE tokens of frames sampled from the context
video. Because the original model does not support starting-frame
conditioning, we replace its original video backbone with the same $14$B
image-to-video generator used in LogiShot. The adapted
model first undergoes supervised warm-up to align with our three-input
format and is then trained using VANS's two-stage Joint-GRPO pipeline. It
generates $49$-frame clips at $480{\times}832$ resolution and $16$~fps
using $50$ denoising steps.

\paragraph{StoryMem~\cite{zhang2025storymem}.}
StoryMem generates sequential shots from a text prompt and a memory bank
of previously seen frames. To adapt it, Qwen2.5-VL~\cite{bai2025qwen25vltechnicalreport} processes the context
video, prompt instruction, and starting frame to produce a target-event
description, while
StoryMem's official semantic selection algorithm with HPSv3 filtering~\cite{ma2025hpsv3}
selects up to three context frames for its memory bank. The starting frame
is provided as the image condition for target-video generation. The adapted
StoryMem model
generates $49$-frame clips at $480{\times}832$ resolution and $16$~fps to
match the other methods.

\subsection{Evaluation Metrics}
\label{sec:supp_metric_details}

We evaluate the generated videos along two primary dimensions, Logical
Correctness (LC) and Visual Consistency (VC), each defined as the
unweighted mean of three sub-metrics. A single VLM judge with a fixed
evaluation prompt and deterministic decoding scores every generated video
on all six sub-metrics using a $1$--$5$ scale.

\paragraph{Logical Correctness.} Three sub-metrics jointly evaluate
whether the generated video depicts the event implied by the inputs,
follows the logical relation, and completes the event.
\emph{Event Match} assesses whether the generated video accurately depicts
this event; a score of $5$ indicates that
the target event is clearly and correctly depicted, $3$ indicates that
the event is partially correct but missing key information, and $1$
signifies that the video depicts an entirely incorrect event.
\emph{Relation Match} evaluates whether the generated event satisfies
the cross-shot relation specified by the prompt instruction; $5$ means the
relation clearly follows from the joint context of the three inputs,
$3$ means the relation is only partially satisfied,
and $1$ denotes that the depicted relation is invalid or unrelated.
\emph{Completion} verifies whether the target event is fully realized,
including its essential action and outcome, rather than merely being
initiated or partially shown; $5$ indicates that the event is completed
and its outcome is clearly visible, $3$ means the event is partially shown but
its outcome is not reached, and $1$ signifies that the event is absent.

\paragraph{Visual Consistency.} Three sub-metrics evaluate whether the
context-relevant participants, their roles, and their states remain
visually and semantically coherent between the context and generated
videos.
\emph{Participant Match} assesses whether the correct context-relevant
people, objects, and tools are present; a score of $5$ indicates that every
relevant participant is clearly present, $3$ means the main participants
are present but a task-relevant one is either missing or hallucinated,
and $1$ signifies that the relevant participants are absent or replaced.
\emph{Role Match} evaluates whether each relevant participant plays the
correct semantic role (e.g., actor, tool, target, or affected object);
$5$ means the role assignments are unambiguously correct, $3$ means some
assignments are ambiguous or incorrect, and $1$ denotes that the role
assignments are absent or entirely incorrect.
\emph{State Match} verifies whether the task-relevant appearances,
attributes, postures, object states, and spatial layouts are compatible
with the context video and the specified relation; $5$ indicates clear
compatibility across appearance and state, $3$ means the relevant
elements are present but a key detail conflicts with the context or
relation, and $1$ signifies that the appearance or state is directly
contradicted or absent.

\paragraph{Score Normalization and Aggregation.} Across all six
sub-metrics, a score of $2$ indicates a generation that is largely
incorrect but retains a recognizable semantic connection, while a score
of $4$ indicates a case that is mostly correct but exhibits minor ambiguity
or drift. Each raw sub-metric score $r\in\{1,\ldots,5\}$ is
normalized to $s=(r-1)/4\in[0,1]$. Sample-level LC and VC are computed as
the means of their respective three normalized sub-metrics. Each sub-metric
reported in Tabs.~\ref{tab:main} and~\ref{tab:ablation} is averaged over the
$800$-sample held-out evaluation set. The two dimensions are not combined
into a single global metric and are reported separately to preserve their
distinct interpretations.

\paragraph{Visual-Quality Metrics.} We additionally report three
visual-quality diagnostics computed using VBench~\cite{huang2024vbench}
implementations on the $800$ outputs generated by each method.
\emph{Subject consistency} (\textbf{Subj.}) averages cosine similarities
between the DINO~\cite{caron2021emerging} features of each noninitial frame
and those of its preceding and first frames. \emph{Motion smoothness}
(\textbf{Mot.}) uses AMT~\cite{li2023amt} to interpolate intermediate
frames from their temporal neighbors and compares the predictions with
the corresponding generated frames. To preserve a paired comparison
with identical sample composition across methods, we compute
\emph{temporal flickering} (\textbf{Tmp.}) by applying VBench's normalized
adjacent-frame pixel-difference score to the same $800$ samples for every
method. Higher temporal-flickering scores indicate less frame-to-frame
variation. The unweighted mean of the three metrics (\emph{Avg.}) is
reported as a compact summary.

\section{Additional Results and Analysis}
\label{sec:supp_human_eval}

Complementing the evaluation in Sec.~\ref{sec:experiments}, this section presents the user study, the validation and residual
analysis of the VLM judge, additional ablations of MCG's dense
multimodal cues, and the qualitative failure analysis.

\subsection{User Study}
\label{sec:supp_user_study}

\begin{figure}[h]
  \centering
  \includegraphics[width=\linewidth]{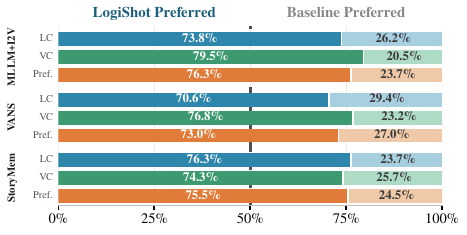}
  \caption{User study win rates. Pairwise A/B win rates for
  LogiShot against each baseline on Logical Correctness (LC),
  Visual Consistency (VC), and overall preference (Pref.).}
  \label{fig:user_study}
\end{figure}

Fig.~\ref{fig:user_study} presents the pairwise A/B win rates from a blind
user study involving $20$ annotators. The compared pairs are drawn from
the same held-out evaluation set. Each annotator evaluated $90$ anonymized video pairs
($30$ comparing LogiShot with each of the three baselines). For each pair,
the annotator provided separate judgments for Logical Correctness, Visual
Consistency, and overall preference, yielding $600$ judgments per baseline
for each criterion. Ties were scored as $0.5$. LogiShot was preferred over
all three baselines across all criteria.

\subsection{VLM Judge Validation and Residual Analysis}
\label{sec:supp_judge_val}
\begin{table*}[t]
  \centering
  \setlength{\tabcolsep}{1mm}
  \renewcommand{\arraystretch}{1.05}
  {\footnotesize
  \begin{tabular}{lccccc}
    \toprule
    Method & $n$ & Human LC~$\uparrow$ [95\%\,CI]
                 & Human VC~$\uparrow$ [95\%\,CI]
                 & $\Delta$LC [95\%\,CI]
                 & $\Delta$VC [95\%\,CI] \\
    \midrule
    MLLM+I2V   & 150 & 0.693~[0.640,\,0.745] & 0.784~[0.738,\,0.825]
                     & $+0.152$~[$+0.091$,\,$+0.213$]
                     & $+0.087$~[$+0.040$,\,$+0.135$] \\
    VANS       & 150 & 0.773~[0.723,\,0.819] & 0.814~[0.771,\,0.854]
                     & $+0.073$~[$+0.013$,\,$+0.134$]
                     & $+0.057$~[$+0.008$,\,$+0.107$] \\
    StoryMem   & 150 & 0.728~[0.670,\,0.784] & 0.825~[0.783,\,0.865]
                     & $+0.117$~[$+0.058$,\,$+0.178$]
                     & $+0.046$~[$+0.000$,\,$+0.093$] \\
    \rowcolor{mygray}
    \textbf{LogiShot}
               & 150 & \textbf{0.845}~[0.805,\,0.883]
                     & \textbf{0.871}~[0.840,\,0.899]
                     & --- & --- \\
    \bottomrule
  \end{tabular}}
  \caption{Per-method mean human scores on the $150$-case validation subset.}
  \label{tab:supp_human_means}
\end{table*}

We validate the VLM judge through three complementary analyses on
a stratified $150$-case validation subset ($30$ per relation type).
For each case, the outputs of all four methods are independently scored by
two blinded annotators on both dimensions, resulting in $600$ evaluation units.
We report (i) rank correlation between the judge and human scores for
each dimension; (ii) per-method human means, providing a judge-independent view
of the performance ordering reported in Tab.~\ref{tab:main}; and (iii)
per-method residuals between the judge scores and the mean human scores, to examine
whether the judge systematically favors any particular method.

Each annotator applied five-point rubrics aligned with the two
evaluation dimensions. Method identities were blinded, and case ordering was
randomized independently for each annotator.
Tab.~\ref{tab:supp_judge_corr} reports Spearman's $\rho$ and
quadratic-weighted Cohen's $\kappa$ for the judge--human comparison on
each dimension, together with the quadratic-weighted inter-annotator
$\kappa$ computed after pooling the paired LC and VC ratings. For
judge--human $\kappa$, the normalized judge score $s$
is mapped back to the five-point scale as $1+4s$. Both this value and the mean
of the two human ratings are then assigned to the nearest integer category
using round-to-nearest with ties to even. Aggregated across all $600$
units, $\rho{=}0.68$ for LC and $\rho{=}0.59$ for VC indicate positive
rank correlations between the judge and human scores. The correlation
is higher for LC than for VC, consistent with the greater subjectivity
of Visual Consistency judgments. Quadratic-weighted Cohen's $\kappa$
further measures category-level agreement, reaching $0.84$ for LC and
$0.76$ for VC. The inter-annotator $\kappa{=}0.81$ indicates strong
consensus between the human raters. In Tab.~\ref{tab:supp_judge_corr},
$n$ denotes the number of (case, method) evaluation units.

\begin{table}[h]
  \centering
  \setlength{\tabcolsep}{1mm}
  \renewcommand{\arraystretch}{1.05}
  {\footnotesize
  \begin{tabular}{lcccccc}
    \toprule
    \multirow{2}{*}{Relation}
      & \multirow{2}{*}{$n$}
      & \multicolumn{2}{c}{$\rho$}
      & \multicolumn{2}{c}{$\kappa_{\text{VLM--hum}}$}
      & \multirow{2}{*}{$\kappa_{\text{inter}}$}\\
    \cmidrule(lr){3-4}\cmidrule(lr){5-6}
      &      & LC   & VC   & LC   & VC   & \\
    \midrule
    Progression       & 120 & 0.66 & 0.65 & 0.81 & 0.76 & 0.79 \\
    Parallel          & 120 & 0.63 & 0.63 & 0.85 & 0.78 & 0.78 \\
    Causal            & 120 & 0.70 & 0.42 & 0.83 & 0.65 & 0.80 \\
    Conditional       & 120 & 0.76 & 0.72 & 0.89 & 0.84 & 0.85 \\
    Overview/Detail   & 120 & 0.65 & 0.49 & 0.81 & 0.69 & 0.79 \\
    \midrule
    All               & 600 & \textbf{0.68} & \textbf{0.59}
                            & \textbf{0.84} & \textbf{0.76}
                            & \textbf{0.81} \\
    \bottomrule
  \end{tabular}}
  \caption{Judge--human correlation and agreement by relation type, reported for Logical Correctness and Visual Consistency.}
  \label{tab:supp_judge_corr}
\end{table}

Tab.~\ref{tab:supp_human_means} reports the per-method mean human score
on this subset alongside $95\%$ bootstrap confidence intervals ($10{,}000$
resamples), as well as the paired LogiShot-minus-baseline gap with its
corresponding CI. We average the two annotators for each case and bootstrap
the paired gaps by case. LogiShot ranks first on both dimensions under
human evaluation. The $95\%$ confidence intervals for all six paired gaps
have nonnegative lower bounds; five are strictly positive, while the sixth
(StoryMem VC) has a lower bound at zero. LogiShot's top ranking is therefore reproduced
under judge-independent human evaluation. Among the
baselines, the ordering reproduces the automated results in
Tab.~\ref{tab:main}, with the exception of a minor swap between StoryMem
and MLLM+I2V on Logical Correctness---a variation that falls well within the
overlapping CIs of the two weakest methods on that axis.
Paired gaps are computed from unrounded case-level differences and
rounded only for display; they may therefore differ slightly from
differences between the displayed means.

Tab.~\ref{tab:supp_judge_residual} reports the mean judge--human residual
for each method, computed per case as the score assigned by the VLM judge minus the mean
score of the two human annotators, together with $95\%$ confidence intervals
obtained through case-level bootstrap resampling. For each method, the judge
and human scores are evaluated on the same $150$ cases. Across methods, the
mean residual ranges from $0.06$ to $0.12$ on Logical Correctness and from
$0.05$ to $0.07$ on Visual Consistency, indicating a consistent positive
offset in the judge scores. LogiShot does not exhibit the largest residual
on either dimension: MLLM+I2V has the largest LC residual at $+0.124$, while
StoryMem has the largest VC residual at $+0.069$. LogiShot's residuals remain
within the range observed for the three baselines, providing no indication
of a method-specific positive offset in favor of LogiShot.

\begin{table}[t]
  \centering
  \setlength{\tabcolsep}{1mm}
  \renewcommand{\arraystretch}{1.05}
  {\footnotesize
  \begin{tabular}{lccc}
    \toprule
    Method & $n$
      & \begin{tabular}[t]{@{}c@{}}LC residual\\\mbox{[95\%\,CI]}\end{tabular}
      & \begin{tabular}[t]{@{}c@{}}VC residual\\\mbox{[95\%\,CI]}\end{tabular} \\
    \midrule
    MLLM+I2V & 150
      & \begin{tabular}[t]{@{}c@{}}$+0.124$\\\mbox{[$+0.101$,\,$+0.149$]}\end{tabular}
      & \begin{tabular}[t]{@{}c@{}}$+0.057$\\\mbox{[$+0.031$,\,$+0.084$]}\end{tabular} \\
    VANS & 150
      & \begin{tabular}[t]{@{}c@{}}$+0.064$\\\mbox{[$+0.043$,\,$+0.085$]}\end{tabular}
      & \begin{tabular}[t]{@{}c@{}}$+0.048$\\\mbox{[$+0.026$,\,$+0.070$]}\end{tabular} \\
    StoryMem & 150
      & \begin{tabular}[t]{@{}c@{}}$+0.061$\\\mbox{[$+0.039$,\,$+0.083$]}\end{tabular}
      & \begin{tabular}[t]{@{}c@{}}$+0.069$\\\mbox{[$+0.045$,\,$+0.093$]}\end{tabular} \\
    \rowcolor{mygray}
    \textbf{LogiShot} & 150
      & \begin{tabular}[t]{@{}c@{}}$+0.064$\\\mbox{[$+0.046$,\,$+0.083$]}\end{tabular}
      & \begin{tabular}[t]{@{}c@{}}$+0.065$\\\mbox{[$+0.042$,\,$+0.089$]}\end{tabular} \\
    \bottomrule
  \end{tabular}}
  \caption{Per-method judge--human residuals on the $150$-case validation subset.}
  \label{tab:supp_judge_residual}
\end{table}

\subsection{Ablations on MCG's Dense Multimodal Cues}
\label{sec:supp_ablation_mcg}

\begin{table}[t]
  \centering
  \setlength{\tabcolsep}{1mm}
  \renewcommand{\arraystretch}{1.05}
  {\footnotesize
  \begin{tabular}{l|ccc|ccc}
    \toprule
    & \multicolumn{3}{c|}{Logical Correctness\,$\uparrow$}
    & \multicolumn{3}{c}{Visual Consistency\,$\uparrow$}\\
    \cmidrule(lr){2-4}\cmidrule(lr){5-7}
    Variant & EvMt & RelMt & Comp
            & PartMt & RoleMt & StMt\\
    \midrule
    Text-only
      & 0.790 & 0.804 & 0.791
      & 0.850 & 0.845 & 0.830 \\
    Matched cues
      & 0.833 & 0.849 & 0.828
      & 0.882 & 0.879 & 0.875 \\
    Shuffled cues
      & 0.771 & 0.789 & 0.766
      & 0.846 & 0.840 & 0.828 \\
    Dense cues only
      & \multicolumn{6}{c}{N/A (no recognizable target event)} \\
    \rowcolor{mygray}
    \textbf{Full}
      & \textbf{0.901} & \textbf{0.939} & \textbf{0.887}
      & \textbf{0.938} & \textbf{0.931} & \textbf{0.931} \\
    \bottomrule
  \end{tabular}}
  \caption{Additional ablations of MCG's dense multimodal cues. The
  variant using only dense cues does not produce recognizable target events, so
  its evaluation metrics are not applicable.}
  \label{tab:supp_ablation_mcg}
\end{table}

\begin{figure}[t]
  \centering
  \includegraphics[width=\linewidth]{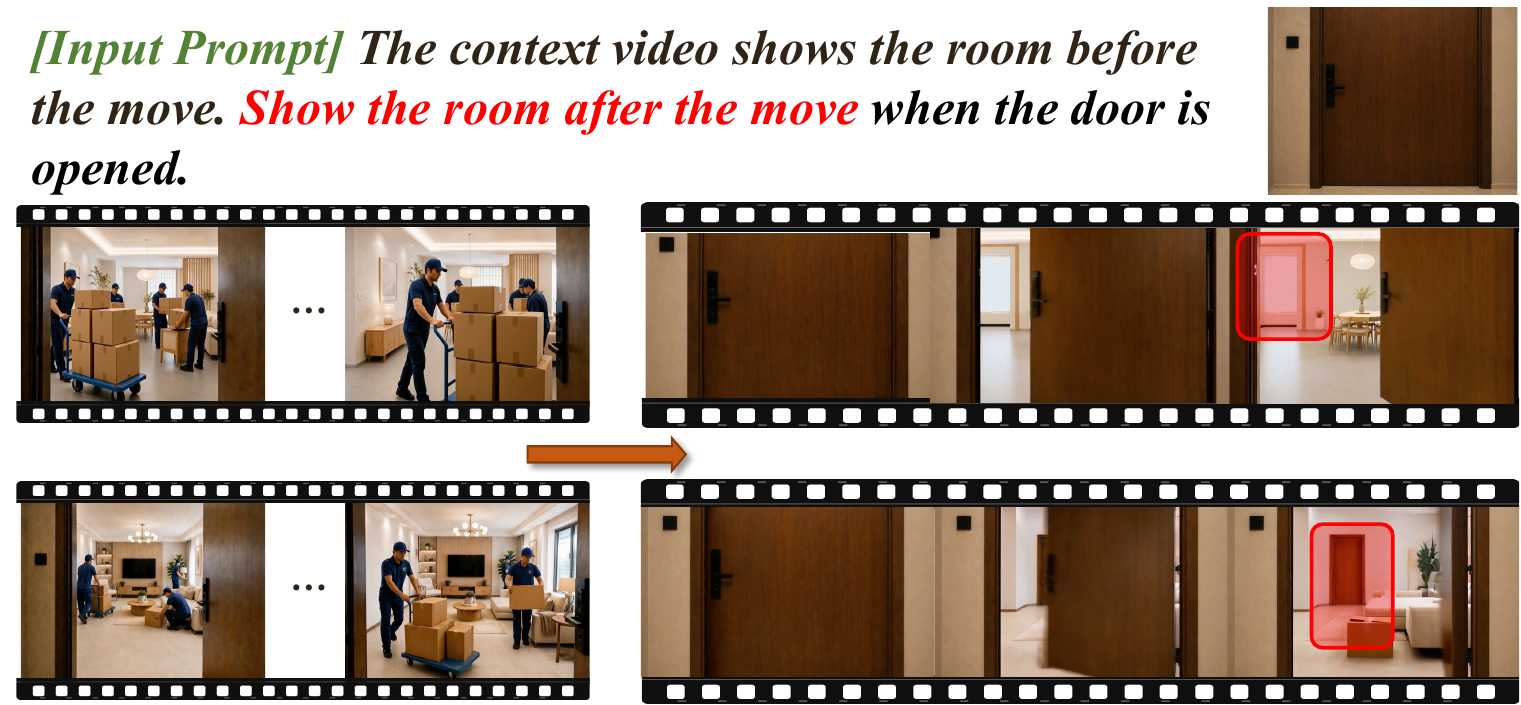}
  \caption{Failure case. The prompt relates the room before and after the move,
  requiring the target video to preserve the wall layout and
  object placement in the newly visible interior. Red boxes mark
  inconsistencies with the context video: an extra door appears, the
  television wall becomes a plain surface, and unrelated boxes remain
  on the floor.}
  \label{fig:bad_case}
\end{figure}

While Tab.~\ref{tab:ablation} evaluates MCG
and VM, we introduce two additional experiments here to specifically
analyze the dense multimodal cues used by MCG. All cue-specific variants
disable VM to isolate MCG. \emph{Text-only} disables both MCG and VM,
retaining the decoded target-event description and starting-frame
conditioning while removing the dense multimodal cues.
\emph{Matched cues} enables MCG with cues derived from the corresponding
$(V,p,I_0)$ while keeping VM disabled. The \emph{Full} model enables
both MCG and VM and is included only as a reference.

First, to test whether
these cues carry context-specific relational information from
$(V,p,I_0)$ rather than serving merely as a generic dense signal, we
replace $P(\mathbf{H})$ with $P(\mathbf{H}^{\text{shuf}})$, where
$\mathbf{H}^{\text{shuf}}$ is computed by the same VLM using an
\emph{unrelated} context video drawn uniformly at random from the
800-sample held-out evaluation set together with the original prompt instruction and
starting frame. The replacement preserves the sequence length and data
type, while the decoded target-event description, starting-frame
conditioning, and sampling protocol remain unchanged.
Tab.~\ref{tab:supp_ablation_mcg} demonstrates that introducing
mismatched dense cues fails to recover the $+0.042$ Logical Correctness
gain produced by matched dense cues. In fact, the
shuffled variant drops $0.020$ below the Text-only variant on Logical
Correctness and $0.004$ below on Visual Consistency. This result
indicates that MCG's dense cues carry context-specific information and
that mismatched cues can degrade generation.

Second, we test whether the dense cues can replace the decoded
target-event description. The \emph{Dense cues only} variant keeps VM
disabled and removes the decoded description while retaining the
matched dense cues and starting-frame conditioning. It fails to
generate evaluable target videos because its outputs do not depict
recognizable target events.
Accordingly, all evaluation metrics for this variant are
marked as N/A in Tab.~\ref{tab:supp_ablation_mcg}. These results suggest
that, under the current conditioning setup, dense cues alone are insufficient
to reliably specify the target event. This finding supports retaining the decoded
target-event description as explicit semantic guidance and using the dense
cues as complementary visual-semantic information.

\subsection{Qualitative Failure Analysis}
\label{sec:supp_failure_modes}

LogiShot can fail when generation requires resolving complex cross-shot
relations and integrating spatially dispersed details from visually noisy
contexts. Fig.~\ref{fig:bad_case} presents one such case. Although the
door-opening action is generated correctly, the target video does not
fully preserve the scene structure from the context video: an extra door
appears on the wall, the television wall becomes a plain surface, and
unrelated boxes remain on the floor. These errors suggest that the combination
of visual clutter and relational complexity can limit scene-level visual
consistency.

\end{document}